%% file: main.tex
\documentclass[10pt,twocolumn,letterpaper]{article}

\usepackage[pagenumbers]{cvpr} 

\input{preamble}
\definecolor{cvprblue}{rgb}{0.21,0.49,0.74}
\usepackage[pagebackref,breaklinks,colorlinks,allcolors=cvprblue]{hyperref}

\usepackage{wrapfig}
\usepackage{array}
\usepackage{xspace}
\usepackage{multirow} 
\usepackage{float}
\usepackage{tikz}
\usetikzlibrary{positioning}
\usepackage{newfloat}
\usepackage{listings}
\usepackage{booktabs}
\newcolumntype{C}[1]{>{\centering\arraybackslash}p{#1}}

\newcommand{\model}{SafeFix\xspace}
\newcommand{\hibug}{HiBug\xspace}
\newcommand{\controlnet}{ControlNet\xspace}
\newcommand{\qwen}{Qwen2.5-VL\xspace}
\newcommand{\llava}{LLaVA-v1.5\xspace}
\usepackage[utf8]{inputenc} 
\usepackage[T1]{fontenc}    
\usepackage{url}            
\usepackage{amsmath}        
\usepackage{amsfonts}       
\usepackage{nicefrac}       
\usepackage{microtype}      
\usepackage{xcolor}         
\usepackage{enumitem}  %
\usepackage{hyperref}
\usepackage{caption}
\def\paperID{21546} 
\def\confName{CVPR}
\def\confYear{2026}

\title{SafeFix: Targeted Model Repair via Controlled Image Generation}

\author{%
  Ouyang Xu \quad Baoming Zhang \quad Ruiyu Mao  \quad Yunhui Guo  \\
  The University of Texas at Dallas \\
  \texttt{\{ouyang.xu, baoming.zhang, ruiyu.mao, yunhui.guo\}@utdallas.edu}
}

\begin{document}
\maketitle
\input{sec/0_abstract}    
\section{Introduction}
Despite extensive training on annotated images and evaluation on static test sets, computer vision models can still fail in unexpected ways~\cite{barbu2019objectnet,gao2023adaptive,leclerc20223db}. One common cause is dataset bias: certain subpopulations are underrepresented in the training data. For instance, facial recognition systems have shown significantly higher error rates for darker-skinned women compared to lighter-skinned men, due to imbalanced representation in training datasets~\cite{buolamwini2018gender}. Therefore, identifying and repairing potential model failures caused by underrepresented semantic subpopulations before deployment is essential to improve the reliability of computer vision systems in real-world applications.

Several recent studies have addressed the identification and repair of model failures caused by underrepresented semantic subpopulations. For example, given a natural language description of a potentially underrepresented semantic subpopulation, AdaVision~\cite{gao2023adaptive} introduces a model debugging pipeline that leverages CLIP~\cite{radford2021learning} to iteratively retrieve relevant images from LAION-5B~\cite{schuhmann2022laion} and evaluate model predictions. \hibug~\cite{chen2023hibug} advances interpretable model debugging by automatically discovering failure slices based on shared visual attributes using a pre-trained vision--language model. DCD~\cite{singla2024data} identifies failure modes by collecting misclassified samples, computing a failure direction for distinguishing attributes, and retrieving similar unlabeled examples to expose systematic weaknesses and underrepresented subpopulations.
The latent-space method~\cite{jain2023latentfailures} discovers failure patterns by training a linear SVM in a  latent space, separating correct from incorrect predictions to yield subpopulation errors without labels.
D3M~\cite{jain2023d3m} enhances overall model performance by removing harmful training samples that disproportionately affect underrepresented groups.

While existing methods can effectively identify model errors caused by underrepresented semantic subpopulations, the subsequent and critical step—model repair—still presents challenges. For instance, AdaVision~\cite{gao2023adaptive} employs a simple fine-tuning strategy to retrain the model on retrieved error-causing images.
However, this approach can degrade performance on images where the model originally performed well. \hibug~\cite{chen2023hibug} addresses model repair by turning attributes describing underrepresented subpopulations into language prompts to synthesize new training images via generative models. However, the synthesized images may suffer from \emph{distribution mismatch} compared to the original training data. Moreover, the repair process can introduce new bugs, as the generated images may fail to preserve attributes unrelated to original model errors.

Building on these observations, we propose \model, a targeted model repair method designed to correct failures caused by underrepresented semantic subpopulations through controlled image generation—without degrading overall performance or introducing new errors. Compared to AdaVision~\cite{gao2023adaptive} and DCD~\cite{singla2024data}, which retrieve samples from another dataset, our method generates new examples grounded in the training set to ensure distributional consistency. Unlike \hibug~\cite{chen2023hibug}, which simply composes failure attributes into language prompts, \model anchors generation on real instances in the training set to preserve unrelated attributes and avoid unintended changes. While the latent-space-based filtering method~\cite{jain2023latentfailures} explores repair via diffusion models, their approach relies solely on caption-based prompting and does not guarantee the semantic accuracy of critical attributes. In contrast, \model mitigates the unreliability of generative models in rendering sensitive attributes—such as skin tone or hair color—by further filtering out semantically inconsistent images using a large vision--language model (LVLM)~\cite{bai2025qwen25vl, liu2023improvedllava}, and the LVLM is verified by human audit. This ensures that the final outputs faithfully reflect the intended attribute change (e.g., generating a sad expression on a darker-skinned woman with red hair, whereas standard diffusion models often mistakenly produce a darker-skinned man with black hair instead), enabling precise and robust repair of underrepresented semantic subpopulation failures. 



\model directly addresses two major challenges in existing model repair strategies: {\bf 1)} ensuring the generated images reflect the true failure attributes, and {\bf 2)} aligning them with the original data distribution. By retraining models on this augmentation set, we reduce errors associated with underrepresented semantic subpopulations, where model failures stem from insufficient training coverage of specific attributes. Our contributions are as follows:
\begin{itemize}
\item We propose a targeted model repair pipeline, \model, which leverages conditional text-to-image generation and LVLM-based filtering to synthesize high-quality, attribute-faithful data for correcting model failures arising from underrepresented subpopulations.

\item We show that our approach mitigates distribution mismatch in generated samples, ensuring that the augmented data remains aligned with the original dataset distribution. This alignment helps correct model failures while maintaining overall performance.
\item We demonstrate that retraining with curated, distribution-consistent synthetic data significantly enhances model robustness in underrepresented scenarios—without introducing new errors.

\end{itemize}

\section{Related Work}
\textbf{Failure Pattern Discovery.} \hibug~\cite{chen2023hibug} proposes a pioneering pipeline that identifies interpretable failure cases in vision models by clustering semantically meaningful attributes, revealing both rare categories and spurious correlations.
HiBug2~\cite{chen2025hibug2} extends this approach with more efficient error-slice discovery and a closed-loop debugging mechanism, improving coherence and coverage of discovered model bugs.
Several follow-up works attempt to strengthen interpretability in model debugging~\cite{adebayo2020debugging, adebayo2022posthoc}, though often limited by static failure patterns or inadequate visual grounding.
Beyond \hibug, MODE~\cite{vendrow2023dataset} introduces a state-differential analysis framework that locates internal model faults and proposes data-driven remedies.
TCAV~\cite{kim2018tcav} further enriches interpretability by testing the model’s sensitivity to high-level concepts and by correcting spurious activations at the concept level. 3DB~\cite{leclerc20223db} complements these efforts by constructing structured attribute spaces over failure modes, allowing discovery of underrepresented attributes through unsupervised analysis of visual model errors.

\noindent\textbf{Targeted Synthetic Augmentation via Diffusion Models.}
Diffusion models such as Stable Diffusion~\cite{rombach2022high} and classifier-free guidance~\cite{ho2022classifierfree} enable controllable, semantically faithful image synthesis. These models have proven useful in debugging~\cite{casper2022robust, fang2024data, huang2024data}, counterfactual visualization~\cite{augustin2022diffusion, boreiko2022sparse}, and training data augmentation~\cite{trabucco2023effective, dunlap2023diversify}, especially in few-shot and fine-grained recognition tasks. Recent work like DiGA~\cite{zhang2024distributionally} shows how editing spurious attributes while preserving class semantics can mitigate bias without requiring new annotations. Others, such as GenMix~\cite{lee2024genmix}, propose mixing synthetic and real images to enhance robustness. These advances highlight how targeted generation can shape training distributions to address model weaknesses. Building on these insights, our method forms a closed-loop debugging pipeline that not only identifies failure cases but also synthesizes and integrates targeted images to improve model performance. Compared to prior augmentation or error discovery pipelines~\cite{fang2024data, huang2024data, chen2023hibug}, our approach is more generalizable and less reliant on predefined attribute sets. We draw inspiration from efforts like StylizedImageNet~\cite{geirhos2019imagenet}, debiasing pipelines~\cite{jin2021background}, and adaptive augmentation methods~\cite{mikolajczyk2023targeted, zhao2022adaptive, wang2024domain}, but focus on semantically controllable generation tailored to discovered bugs. 

\noindent\textbf{Multimodal Filtering via LVLMs.} LVLMs like Flamingo~\cite{alayrac2022flamingo} have demonstrated strong capabilities in semantically grounding visual concepts. Recent studies show using LVLMs as filters to select high-quality image-text pairs enhanced dataset quality for downstream tasks~\cite{wang2024finetuned, li2024self}. These approaches outperform traditional methods like CLIP-based filtering by providing fine-grained, attribute-aware analysis of generated samples. In line with these trends, we adopt an LVLM as an automated filtering component, grounding our approach in established methods that leverage LVLMs for semantic validation of generated data. 

\noindent\textbf{Targeted Repair for Rare-Case Bugs.}  
Recent work explores how underrepresented subpopulations induce systematic errors in vision models and how targeted interventions can mitigate such failure modes. DOMINO~\cite{eyuboglu2022rare} discovers coherent failure slices by clustering model errors in a cross‑modal embedding space, enabling automated identification of rare-case bugs without manual slice definitions.
REAL~\cite{parashar2024neglected} focuses on rare visual concepts that are neglected in large-scale vision--language datasets, augmenting them by retrieving semantically similar examples and fine-tuning lightweight classifiers on the retrieved subsets to improve model robustness on these rare categories.
This work reflects a broader shift toward targeted model repair using interpretable diagnostics and subpopulation-aware interventions, aligning with our controlled synthesis and refinement approach.

\begin{figure*}[t]
  \setlength{\dbltextfloatsep}{5pt} 
  \setlength{\abovecaptionskip}{5pt}
  \centering
  \includegraphics[trim={0 6pt 0 1pt}, clip, width=0.84\textwidth]{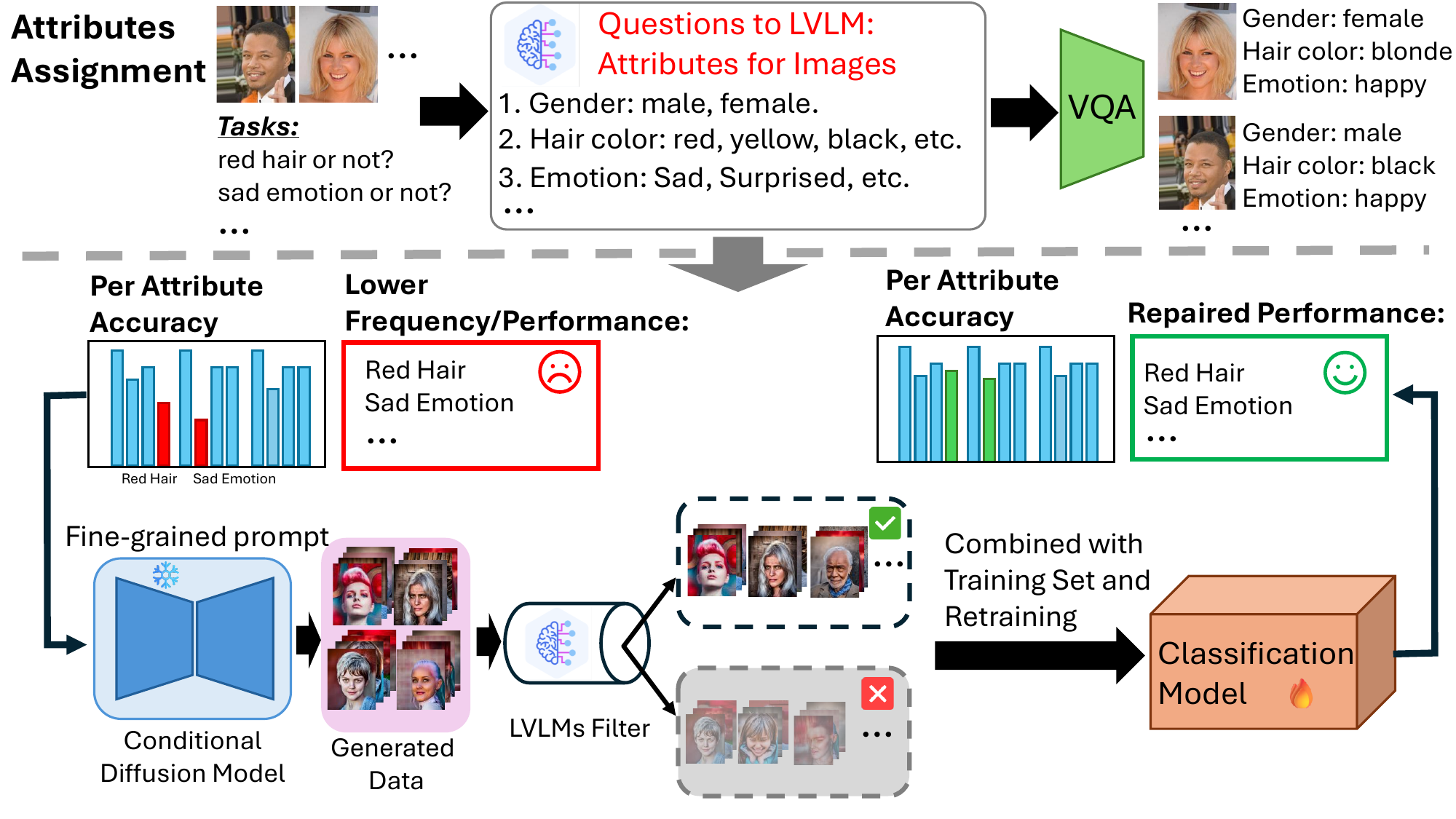}
  \caption{\textbf{Overview of \model}. We propose a targeted model repair pipeline that identifies rare-case failures, generates attribute-specific synthetic images using a conditional diffusion model, filters them via a large vision--language model, and retrains the model to improve accuracy and fix rare-case bugs.}
  \label{fig:overview}
  \vspace{-15pt}
\end{figure*}
\section{Background}

\paragraph{Attribute-based Model Debugging.}  
Let \(x \in \mathcal{X}\) be an input image with its corresponding ground-truth label \(y(x) \in \mathcal{Y}\). A computer vision model \(f_\theta \colon \mathcal{X} \to \mathcal{Y}\) produces a prediction \(f_\theta(x)\) for each input image \(x\). Denote the training, validation, and test splits by \(\mathcal{D}_{\mathrm{train}}\), \(\mathcal{D}_{\mathrm{val}}\), and \(\mathcal{D}_{\mathrm{test}}\), respectively. The full dataset is then given by \(\mathcal{D} = \mathcal{D}_{\mathrm{train}} \cup \mathcal{D}_{\mathrm{val}} \cup \mathcal{D}_{\mathrm{test}}\). The overall validation accuracy is computed as: 
\vspace{-5pt}
\begin{equation}
\mathrm{Acc}(\mathcal{D}_{\mathrm{val}})
=\frac{1}{|\mathcal{D}_{\mathrm{val}}|}
\sum_{(x, y) \in \mathcal{D}_{\mathrm{val}}}
\mathbf{1}\{f_\theta(x) = y(x)\}.
\end{equation}
In attribute-based model debugging~\cite{chen2023hibug}, each image is assigned several attributes that represent different subpopulations. 
In particular, let \(\mathcal{A} = \{a_1, \dots, a_m\}\) be a set of attribute functions, where each \(a_i \colon \mathcal{X} \to \mathcal{V}_i\) maps an image \(x\) to a discrete value \(v \in \mathcal{V}_i\) and $\mathcal{V}_i$ represents the set of all possible values for attribute $i$. A slice $S$ is defined as the set of images that satisfy a conjunction of attribute-value assignments: 
\vspace{-5pt}
\begin{equation}
S = \{ x \mid a_{j_1}(x) = v_{j_1}, \dots, a_{j_k}(x) = v_{j_k}\},
\end{equation}
where \(\mathcal{J} = \{j_1, \dots, j_k\} \subseteq [m]\) is the index set of attributes involved in the slice \(S\).  
The slice accuracy is
\vspace{-5pt}
\begin{equation}
\mathrm{Acc}(S) = \frac{1}{|S|}\sum_{x\in S}\mathbf{1}\{f_\theta(x) = y(x)\}.
\vspace{-5pt}
\end{equation}

\vspace{-5pt}
\paragraph{Underrepresented Semantic Subpopulations.} Rare-case bugs, often caused by underrepresented semantic subpopulations~\cite{eyuboglu2022rare}, are attribute-based failure modes for which the model’s error rate on validation samples matching a target description significantly exceeds its overall error and those samples appear infrequently in the training dataset. We say a slice \(S_r\) is a \emph{rare-case slice} if it constitutes less than a fraction \(\rho\) of the training data, i.e., 
\begin{equation}
|S_r| < \rho \,\bigl|\mathcal{D}_{\mathrm{train}}\bigr|,
\label{eq:rare}
\end{equation}
where \(\rho\) is a \emph{rare threshold} (e.g., 0.05). We flag a candidate slice \(S_r\) as a bug slice \(S_e\) if  
\begin{equation}
\mathrm{Acc}(S_{e}) < \mathrm{Acc}(\mathcal{D}_{\mathrm{val}}) - \epsilon.
\label{eq:low_acc_and_rare}
\end{equation}
where \(\epsilon\) is an \emph{accuracy difference threshold}. Thus, the model shows significantly lower accuracy on the bug slice—which represents an underrepresented semantic subpopulation—relative to its average validation accuracy. A bug slice can be converted to a human-readable bug description (e.g., “people with red hair smiling tend to have low accuracy on the ‘wearing lipstick’ classification task”).

\section{\model}
To address rare-case bugs caused by underrepresented semantic subpopulations, we propose a targeted model repair strategy that leverages controlled image generation and semantic filtering enabled by a large vision--language model. Our goal is to generate synthetic images that accurately represent underrepresented semantic subpopulations while ensuring these images remain aligned with the training distribution. The overall workflow is summarized in Figure~\ref{fig:overview}.

We begin by generating visually controlled images using a text-to-image diffusion model, \textit{Stable Diffusion}, guided by a structure-conditioned controller, \controlnet{}~\cite{zhang2023adding}. Instead of relying solely on language prompts, we generate images conditioned on training data while modifying specific attributes to reflect failure slice semantics. Next, we employ an LVLM to automatically verify whether each generated image accurately reflects the intended attributes. This filtering step ensures that only semantically faithful samples are retained. Finally, we augment the original training dataset with the validated images and retrain the model. \model enhances performance on error-prone regions without compromising overall accuracy or introducing new bugs.

\subsection{Model Diagnosis for Identifying Rare-Case Bugs}
We begin by training a standard computer vision model \(f_\theta\) on the original training set \(\mathcal{D}_{\mathrm{train}}\). After obtaining predictions on the validation set \(\mathcal{D}_{\mathrm{val}}\), similar to \hibug{}, we use a large vision-language model (LVLM; e.g., GPT-4 with vision~\cite{openai2023gpt4}) to propose candidate attributes and a VQA model (e.g., BLIP~\cite{li2022blip}) to assign attribute values \(a_i(x)\) to each image. We extract a set of rare-case bug slices \(\{S_e\}\), each defined by a conjunction of attribute conditions that exhibit both high error rates and low coverage in the dataset. Specifically, to identify underrepresented and error-prone subpopulations, we analyze each attribute \(a_i\) and its associated values \(v \in \mathcal{V}_i\). For each value \(v\), we examine the slice \(S = \{x \mid a_i(x) = v\}\) and compute two quantities: (1) its proportion in the training set, and (2) the model’s accuracy on the corresponding validation samples.

\noindent We flag the slice \(S\) as a rare-case bug if:
\begin{itemize}
    \item The training support \(|S \cap \mathcal{D}_{\mathrm{train}}|\) is below the threshold \(\rho \cdot |\mathcal{D}_{\mathrm{train}}|\).
    \item The validation accuracy \(\mathrm{Acc}(S)\) is significantly below the overall validation accuracy \(\mathrm{Acc}(\mathcal{D}_{\mathrm{val}})\) (by a margin \(\epsilon\)).
\end{itemize}
For example, consider the attribute \textit{hair color}. Suppose only 3\% of the dataset have \textit{red hair}, and the model performs poorly on this group (e.g., 60\% accuracy versus 85\% overall). This makes \textit{red hair} a rare-case bug. In contrast, if \textit{yellow hair} is infrequent but achieves high accuracy, it is not considered a bug. This analysis helps isolate specific attribute values that contribute to systematic model failures.

\subsection{Targeted Generation with Conditional Diffusion Models (CDMs)}

Next, we aim to produce attribute-preserving edits on each original image \(x\), focusing on targeted attributes identified in problematic slices from the previous diagnostic stage. Specifically, we generate visually controlled synthetic images \(x'\) using a text-to-image diffusion model, \textit{Stable Diffusion}, guided by the structure-conditioned controller \controlnet{}~\cite{zhang2023adding}. 

To determine which attribute–value pairs to edit, we first identify rare-case slices \(S_e\) by computing slice support and validation accuracy across attribute conjunctions, as defined in Eq.~\eqref{eq:rare} and Eq.~\eqref{eq:low_acc_and_rare}. Each \(S_e\) contains one or more attribute–value pairs \(\{(a_j, v_j')\}_{j \in \mathcal{J}}\) that are both infrequent and underperforming. These attributes yield significantly lower accuracy than average on the validation set. 
We rank all such attributes by validation error and select the top-\(k\) attributes for augmentation. For each original image \(x\) that does not satisfy all conditions in \(S_e\), we construct an edited variant \(x'\) by modifying the selected attributes \(\{a_j\}_{j \in \mathcal{J}}\) to match the error-prone configuration defined by \(S_e\), while preserving all other visual characteristics and keeping the original label unchanged, i.e., \(y(x') = y(x)\).

Given an original image \(x\) in the training set with attribute assignment \(\{(a_i(x) =  v_i)\}_{i=1}^m\), we construct a modified image \(x'\) by replacing a subset \(\{(a_j, v_j)\}_{j \in \mathcal{J}}\) with \(\{(a_j, v_j')\}_{j \in \mathcal{J}}\), where \(\mathcal{J} \subseteq [m]\) indexes attributes satisfying the slice condition \(S_e\). In practice, this subset is small (i.e., \(|\mathcal{J}| \ll m\)), and the remaining attribute assignments \(\{(a_i, v_i)\}_{i \in [m] \setminus \mathcal{J}}\) are left unchanged. 
For example, suppose \(x\) has attribute assignment \((\textit{black hair}, \textit{happy emotion})\) and label “not wearing lipstick.” If both attributes are part of the rare-case slice \(S_e\), then the synthetic variant \(x'\) is generated with attributes \((\textit{red hair}, \textit{sad emotion})\) while retaining the same label “not wearing lipstick.” This attribute-preserving image generation aligns with fairness-driven augmentation in attribute classification, where rare attributes (e.g., hair color or emotion) are modified while the primary label is held fixed. By retraining the model on these synthetically augmented samples \(\{x'\}\), we encourage the model to correct rare-case bugs without introducing new bugs.




\subsection{Filtering via Large Vision--Language Models}

We found that synthetic images generated by the Conditional Diffusion Model (CDM) can sometimes fail to accurately reflect the intended attribute modifications. To address this, we employ large vision--language models (LVLMs), \qwen{}-7B~\cite{bai2025qwen25vl} and \llava{}-7B~\cite{liu2023improvedllava}, to automatically verify that each generated image correctly exhibits the desired attributes and retains the original label.

Specifically, for each synthetic image \(x'\) generated to satisfy a bug slice \(S_e\), which contains the error attribute–value pairs responsible for bugs, we iterate over each edited pair \((a_j, v_j')\) for \(j = 1, \dots, k\), and query the LVLM with: 
\vspace{-3pt}
\begin{quote}
\centering
\setlength{\parskip}{0pt}
\setlength{\parsep}{0pt}
\texttt{``Does the object have attribute \(a_j\) equal to \(v_j'\)?''} \newline
\texttt{``Is the object in this picture labeled \(y(x)\)?''}
\end{quote}
\vspace{-5pt}

We retain only those images for which the LVLM answers “yes” to all queries and confirms that the label matches the original ground truth \(y(x)\). Thus, after the LVLM filtering, the generated image \(x'\) satisfies the desired attributes and preserves the original label, i.e., \(y(x') = y(x)\). These validated samples are then added to \(\mathcal{D}_{\mathrm{train}}\) for model retraining. We conduct human-audit experiments to verify that the LVLMs can reliably filter out low-quality generated samples.

\subsection{Combining Generated Images with the Original Dataset and Retraining}
To repair the rare-case bugs while maintaining the model's performance on the overall training distribution ~\cite{lee2024genmix}, we augment the training set by adding the validated synthetic images \(\{x'\}\), yielding an updated training set 
$
\mathcal{D}_{\mathrm{train}}' = \mathcal{D}_{\mathrm{train}} \cup \{x'\}.
$
We then retrain the vision model \(f_\theta\) on \(\mathcal{D}_{\mathrm{train}}'\) and evaluate its performance on \(\mathcal{D}_{\mathrm{val}}\). We report both the overall accuracy improvement and the reduction in failure rates(fix rate) on critical rare‐attribute slices \(S_e\), before and after augmentation.

\section{Results}

\begin{table*}[!t]
\centering
\small
\setlength{\tabcolsep}{4pt}
\renewcommand{\arraystretch}{0.95}

\resizebox{1\linewidth}{!}{
\begin{tabular}{l | C{1.44cm} C{1.44cm} C{1.44cm} | C{1.44cm} C{1.44cm} C{1.44cm} | C{1.44cm} C{1.44cm} C{1.44cm}}
\toprule
\multirow{2}{*}{Method} 
  & \multicolumn{3}{c|}{\textbf{ResNet} (base acc: 90.57\%)} 
  & \multicolumn{3}{c|}{\textbf{ViT} (base acc: 85.02\%)} 
  & \multicolumn{3}{c}{\textbf{CLIP} (base acc: 88.32\%)} \\
\cmidrule(lr){2-4} \cmidrule(lr){5-7} \cmidrule(lr){8-10}
  & 1k Images & 5k Images & 10k Images
  & 1k Images & 5k Images & 10k Images
  & 1k Images & 5k Images & 10k Images \\
\midrule
DA-CDM              & 4.03 & 3.29 & 3.08 & 4.07 & 10.21 & 6.81 & 15.32 & 16.52 & 15.58 \\
Mask-ControlNet     & 5.41 & 7.32 & 6.15 & 11.28 & 10.48 & 12.88 & 16.10 & 15.92 & 16.78 \\
HiBug\_Class        & 5.09 & 5.83 & 3.40 & 5.14 & 7.74 & 7.21 & 14.64 & 14.98 & 13.96 \\
HiBug\_Task         & 6.79 & 4.88 & 3.92 & 18.69 & 11.75 & 7.74 & 15.75 & 15.15 & 14.64 \\
\cmidrule(lr){1-10}
\textbf{Ours (L)} & 10.39 & 11.45 & 11.66 & 18.09 & \textbf{18.42} & 15.35 & 22.26 & 22.52 & \textbf{21.75} \\
\textbf{Ours (Q)} & \textbf{14.32} & \textbf{12.09} & \textbf{14.95} & \textbf{19.29} & 15.42 & \textbf{15.95} & \textbf{22.77} & \textbf{23.12} & 20.46 \\
\bottomrule
\end{tabular}
}
\caption{Fix Rate (FR\%) on CelebA for varying numbers of added images, models, and methods. Ours (L) and Ours (Q) denote \llava{}-7B and \qwen{}-7B as the large vision–language model filters, respectively. The highest FR in each column is marked in bold.} 
\label{tab:fr_celeba} 
\end{table*}

\begin{table*}[!t]
\centering
\small
\setlength{\tabcolsep}{4pt}
\renewcommand{\arraystretch}{0.95}

\resizebox{1\linewidth}{!}{
\begin{tabular}{l | C{1.44cm} C{1.44cm} C{1.44cm} | C{1.44cm} C{1.44cm} C{1.44cm} | C{1.44cm} C{1.44cm} C{1.44cm}}
\toprule
\multirow{2}{*}{Method} 
  & \multicolumn{3}{c|}{\textbf{ResNet} (base acc: 71.73\%)} 
  & \multicolumn{3}{c|}{\textbf{ViT} (base acc: 97.42\%)} 
  & \multicolumn{3}{c}{\textbf{CLIP} (base acc: 93.78\%)} \\
\cmidrule(lr){2-4} \cmidrule(lr){5-7} \cmidrule(lr){8-10}
  & 100 Images & 500 Images & 1k Images
  & 100 Images & 500 Images & 1k Images
  & 100 Images & 500 Images & 1k Images \\
\midrule
DA-CDM              & 5.24 & 7.32 & 8.53 & 6.20 & 13.18 & 13.57 & 2.89 & 4.34 & 5.95 \\
Mask-ControlNet     & 1.73 & 2.87 & 2.33 & -6.59 & 10.47 & 3.88 & -5.47 & 1.61 & -1.29 \\
HiBug\_Class        & 1.80 & 1.49 & 0.74 & -11.63 & -13.57 & -5.43 & 2.25 & 6.91 & 0.80 \\
HiBug\_Task         & 6.30 & 4.07 & 5.77 & 9.30 & 13.95 & 6.20 & -7.88 & 6.43 & 5.95 \\
\cmidrule(lr){1-10}
\textbf{Ours (L)}   & 8.38 & \textbf{8.70} & 7.75 & 30.62 & \textbf{29.07} & 26.74 & 11.41 & 16.40 & 10.61 \\
\textbf{Ours (Q)}   & \textbf{9.13} & 6.01 & \textbf{11.14} & \textbf{31.01} & 25.97 & \textbf{38.37} & \textbf{12.70} & \textbf{17.68} & \textbf{18.81} \\
\bottomrule
\end{tabular}
}
\caption{Fix Rate (FR \%) on ImageNet10 under varying image counts, models, and methods. }
\label{tab:fr_imagenet10}

\end{table*}

\subsection{Experimental Setup}
\textbf{Datasets.}  
We evaluate our method on two classification tasks that exhibit attribute-based failure modes and we use an 8:1:1 split for training, validation, and testing, respectively.

\textit{Lipstick-wearing classification.} 
We use the CelebA dataset~\cite{liu2015faceattributes} and follow the same data split protocol as~\cite{chen2023hibug} (80{,}000, 10{,}000, 10{,}000 for train/val/test). The task is to predict whether a person is wearing lipstick, a label known to be correlated with other attributes such as gender and hair color. In the following experiments, we refer to this dataset simply as \emph{CelebA}.

\textit{ImageNet-10 classification.}  
We construct a 10-class subset of ImageNet~\cite{deng2009imagenet} containing the following categories:  
\texttt{backpack}, \texttt{barber chair}, \texttt{coffee mug}, \texttt{desk}, \texttt{electric guitar}, \texttt{park bench}, \texttt{pitcher}, \texttt{purse}, \texttt{rocking chair}, and \texttt{water bottle}. Each class contains 1,300 images. This subset is selected to study classification failures related to visual attributes such as texture and color. In the following experiments, we refer to this dataset simply as \emph{ImageNet10}.

\noindent\textbf{Baselines.} We compare our method with four recent baselines that use generative augmentation strategies:

\begin{itemize}
\item \textbf{DA-CDM ~\cite{fang2024data}} is a data augmentation method for object detection. It uses a controllable diffusion model guided by visual priors from original images, which enables direct reuse of existing bounding box annotations. It then applies a category-calibrated CLIP score to filter generated data and ensure high-quality, text-aligned samples.
\item \textbf{Mask-ControlNet ~\cite{huang2024data}} is a text-guided image generation pipeline that uses ControlNet with facial occlusion masks to synthesize diverse face images under specific occlusions as a data augmentation method to improve model robustness.

    \item \textbf{HiBug\_Class ~\cite{chen2023hibug}.}  
    A \underline{\textbf{Class}}-level method that augments training data with synthetic images. For each class, it uses a diffusion model with a prompt:

    \textit{``A photo of (*label).''}

    \item \textbf{HiBug\_Task~\cite{chen2023hibug}.}  
    A \underline{\textbf{Task}}-level variant of HiBug\_Class that targets failure-prone attributes. It selects attribute slices with the highest validation error and generates prompts to guide a diffusion model:

    - \textit{CelebA}:  
    \textit{``A photo of a \{gender\} \{beard clause\} \{makeup clause\} \{lipstick clause\} (*label), with \{hair\} hair and \{skin\} skin, looking \{emotion\}, appearing \{age\}.''}

    - \textit{ImageNet10}:  
    \textit{``A photo of a \{color\} \{class name\} (*label) with \{texture\} texture, located \{object position\}, appearing \{object size\}, in a \{background\}, under \{lighting\} lighting, during \{time\}, from a \{perspective\} perspective.''}

    \textit{Note:} Unless otherwise specified, ``\hibug'' refers to this optimized ``HiBug\_Task'' variant in the paper. We do not compare with HiBug2~\cite{chen2025hibug2}, which is a data selection method and differs from the synthetic generation strategy.

\end{itemize}

\noindent\textbf{Metrics.} We focus on \textbf{targeted improvements for rare slices}, which are critical for fairness and safety, while maximizing overall classification accuracy. Following prior work~\cite{lai2023improvingpredictionbackwardcompatiblilitynlp}, we report the \textbf{\emph{Fix Rate} (FR)}, defined as
\vspace{-5pt}
\begin{equation}
FR = \frac{Acc_{\text{after}} - Acc_{\text{before}}}{1 - Acc_{\text{before}}},
\end{equation}
which measures the fraction of previously misclassified samples that are corrected. Here, $Acc_{\text{after}}$ denotes the accuracy obtained after applying the proposed method, and $Acc_{\text{before}}$ represents the baseline accuracy from standard training using the vision model.

\noindent \textbf{Implementation Details.} All experiments use an NVIDIA A100 GPU.  
We take the CelebA ``wearing lipstick'' classification task as an example.  
We evaluate three backbone architectures for this classification task: ResNet-18~\cite{he2016deep}, ViT-B/16~\cite{dosovitskiy2020vit}, and CLIP (ViT-B/32)~\cite{radford2021learning}, all of which are initialized with random weights. All vision models are trained using cross-entropy loss. For rare-case bug discovery, we set the rarity threshold \(\rho = 0.05\) and the accuracy difference threshold \(\epsilon = 0.03\). For synthetic augmentation, we use \controlnet~\cite{zhang2023adding}, a CDM based on Stable Diffusion 1.5~\cite{rombach2022high} and conditioned on soft HED boundaries~\cite{xie2015holistically}, with 30 DDIM inference steps. Soft HED boundaries preserve structural details, making this approach suitable for attribute-preserving edits like recoloring and stylizing.
To filter generated images, we employ the LVLMs \qwen{}-7B~\cite{bai2025qwen25vl} and \llava{}-7B~\cite{liu2023improvedllava}. Generating 1,000 images with ControlNet takes about one hour, and filtering these 1,000 images with the LVLM takes ten minutes. Filtering accuracy for most attributes (e.g., hair color, skin tone) exceeds 90\%, which is consistent with the human audit in Section~\ref{sec:human_audit}. Combining three attributes yields at least a 70\% pass rate for filtered images, showing that the diffusion model produces high-quality samples and that the LVLM filtering is reliable.

\subsection{Main Results} 
\label{sec:main_results}

We summarize the main results on the CelebA and ImageNet10 datasets in Tables~\ref{tab:fr_celeba} and~\ref{tab:fr_imagenet10}, respectively.

On \textbf{CelebA}, we select rare-case bugs defined by attribute–value combinations \texttt{red hair}, \texttt{brown skin}, and \texttt{sad emotion} for ResNet and ViT, and \texttt{yellow hair}, \texttt{brown skin}, and \texttt{sad emotion} for CLIP, based on the most frequent patterns identified among failure slices. \model{} consistently achieves the highest test accuracy, i.e., the highest FR, across all models (ResNet, ViT, and CLIP) and varying levels of synthetic augmentation. For example, with 1{,}000 added images, our method improves FR by +14.32\% (ResNet), +10.29\% (ViT), and +22.77\% (CLIP) relative to their base accuracies. These results show that our attribute-targeted augmentation and filtering pipeline is effective in repairing rare-case failure slices, outperforming both CDM-based and \hibug{} baselines.

On \textbf{ImageNet10}, similar trends emerge, as shown in Table~\ref{tab:fr_imagenet10}. For ResNet and ViT, we target rare-case bugs involving \texttt{pink color} and \texttt{fabric texture}, while for CLIP we use \texttt{orange color} and \texttt{fabric texture}. Across all models, our proposed method consistently surpasses the baseline methods. Specifically, our method improves ResNet accuracy by +11.14\% (with 1{,}000 images) compared to the base model. ViT and CLIP also exhibit a steady improvement compared to other methods. Table \ref{tab:imagenet_resnet_only} shows that the proposed SafeFix also achieves better overall accuracy on \textbf{ImageNet10} compared with the baselines.



\begin{table}
\centering
\small
\setlength{\tabcolsep}{4pt}
\renewcommand{\arraystretch}{0.95}

\resizebox{0.7\linewidth}{!}{
\begin{tabular}{l | C{1.44cm} C{1.44cm} C{1.44cm}}
\toprule
\multirow{2}{*}{Method} 
  & \multicolumn{3}{c}{\textbf{ResNet}} \\
\cmidrule(lr){2-4}
  & 100 Images & 500 Images & 1k Images \\
\midrule
Base               & \multicolumn{3}{c}{71.73} \\
\cmidrule(lr){1-4}
DA-CDM             & 73.21 & 73.80 & 74.14 \\
Mask-ControlNet    & 72.22 & 72.54 & 72.39 \\
HiBug\_Class       & 72.24 & 72.15 & 71.94 \\
HiBug\_Task        & 73.51 & 72.88 & 73.36 \\
\cmidrule(lr){1-4}
\textbf{Ours (L)}  & 74.10 & \textbf{74.19} & 73.92 \\
\textbf{Ours (Q)}  & \textbf{74.31} & 73.43 & \textbf{74.88} \\
\bottomrule
\end{tabular}
}
\caption{Test accuracy (\%) on ImageNet10 using ResNet-18.}
\label{tab:imagenet_resnet_only}
\vspace{-14pt}
\end{table}

\begin{figure}
  \centering
  \begin{subfigure}[b]{0.32\linewidth}
    \includegraphics[width=\linewidth]{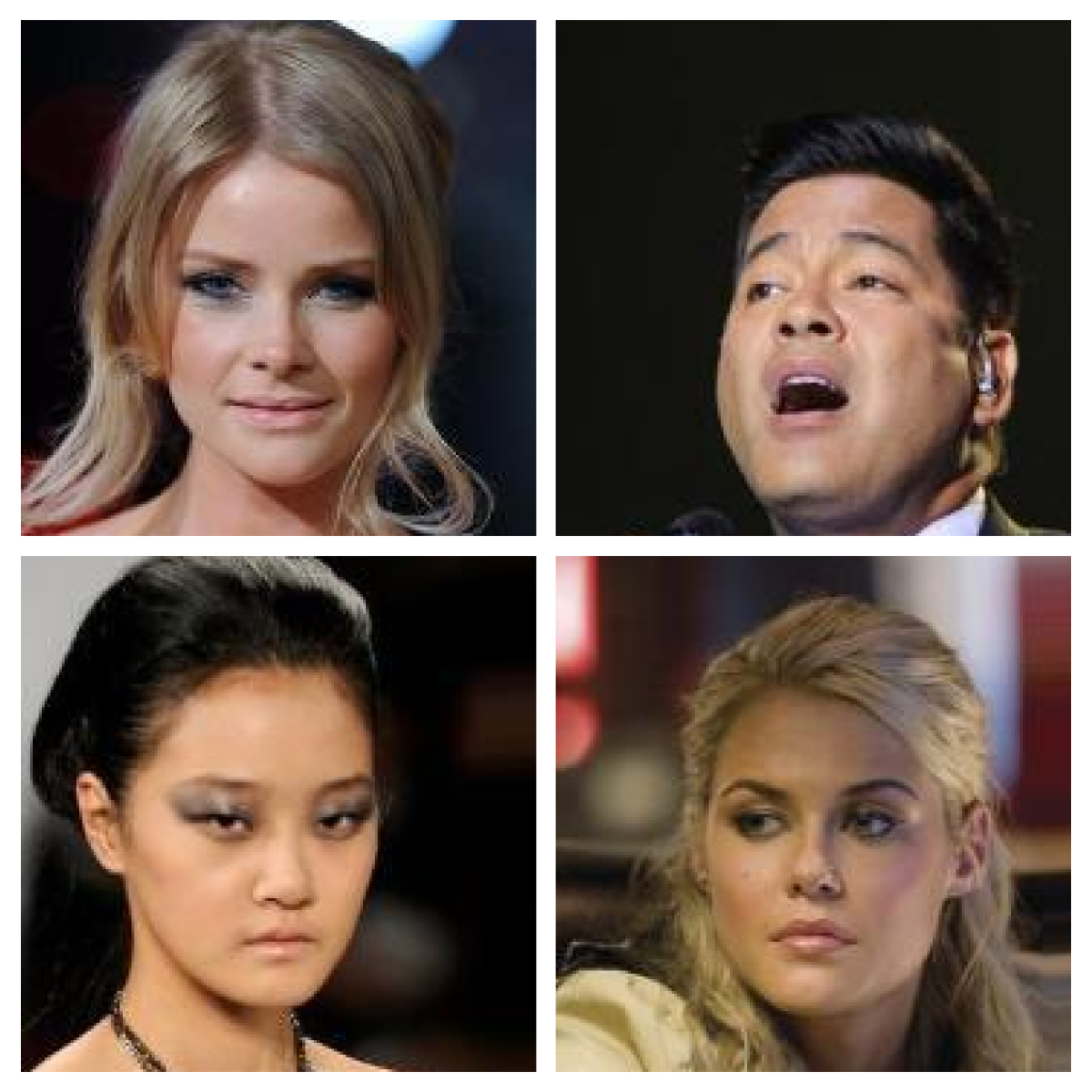}
    \caption{Original Images}
  \end{subfigure}
  \hfill
  \begin{subfigure}[b]{0.32\linewidth}
    \includegraphics[width=\linewidth]{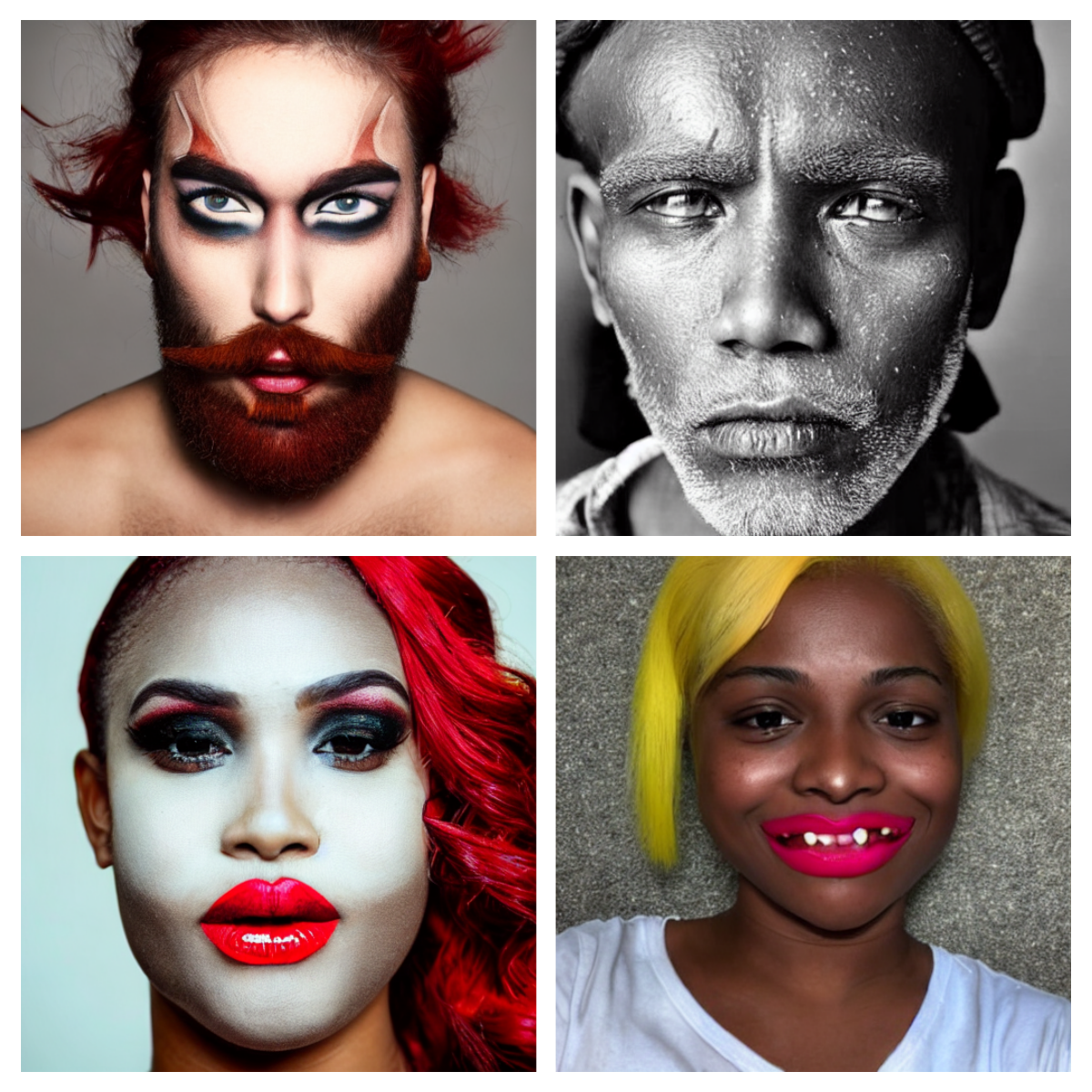}
    \caption{HiBug-Generated}
  \end{subfigure}
  \hfill
  \begin{subfigure}[b]{0.32\linewidth}
    \includegraphics[width=\linewidth]{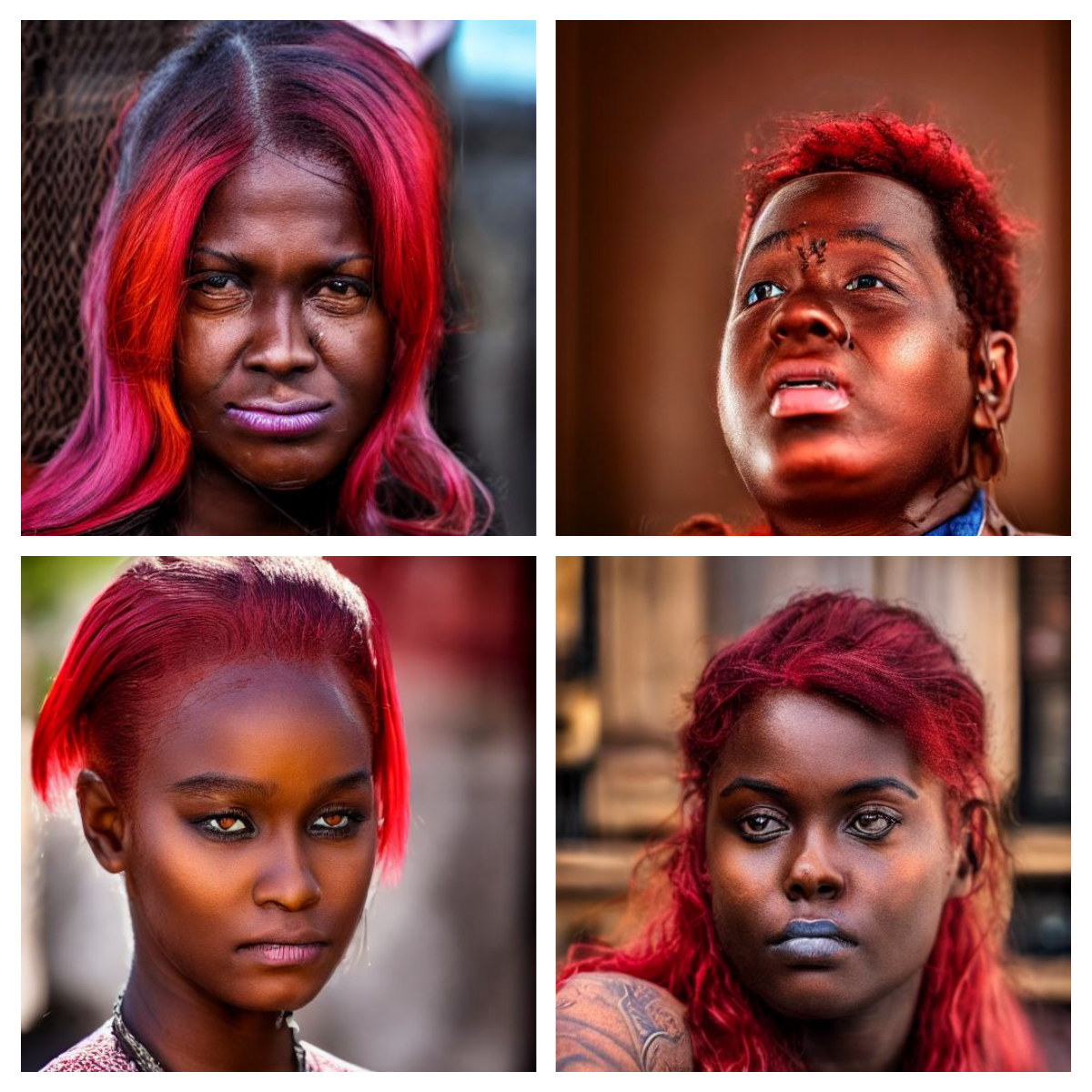}
    \caption{SafeFix-Generated}
  \end{subfigure}
  \vspace{-4pt}
  \caption{Comparison of generated images from different methods with edited attributes \texttt{red hair}, \texttt{brown skin}, and \texttt{sad emotion}. \hibug{} often produces invalid or imprecise samples due to the lack of conditional generation and semantic filtering. In contrast, \model{} generates attribute-faithful images that specifically target rare-case bugs.
}
  \label{fig:picture_compare}
\end{figure}

\noindent\textbf{Analysis.}
Combined with Tables~\ref{tab:fr_celeba}, \ref{tab:fr_imagenet10}, \ref{tab:imagenet_resnet_only}, and \emph{all other test-accuracy results} in Appendix~\ref{sec:appendix_acc}, our experiments show that baseline performance is often unstable. For instance, baselines like DA-CDM and Mask-ControlNet show some improvement on object and facial attribute tasks, respectively, due to their use of conditional diffusion models. However, they are \textbf{not targeted model repair methods}, so their overall performance is inferior to both HiBug and \model. This lack of a targeted strategy means their gains are task-specific and \textbf{not robustly transferable}.
Similarly, the accuracy of \hibug{} does not improve substantially. This is likely because it does not precisely target failure-critical attributes, instead applying general augmentations that can introduce noise or modify unintended features. As shown in Figure~\ref{fig:picture_compare}, this issue is compounded by the lack of LVLM filtering, resulting in generated data that is often misaligned with the intended rare-case fixes.

In contrast, \textbf{our method (\model) consistently shows stable or improving performance}. This robustness indicates our attribute-targeted augmentation and LVLM filtering are highly effective at correcting failure-prone subpopulations with meaningful data. 


\subsection{\model Can Effectively Fix Rare-case Bugs}

\begin{figure}[t]
  \centering
  \begin{subfigure}[b]{0.49\linewidth}
    \includegraphics[width=\linewidth]{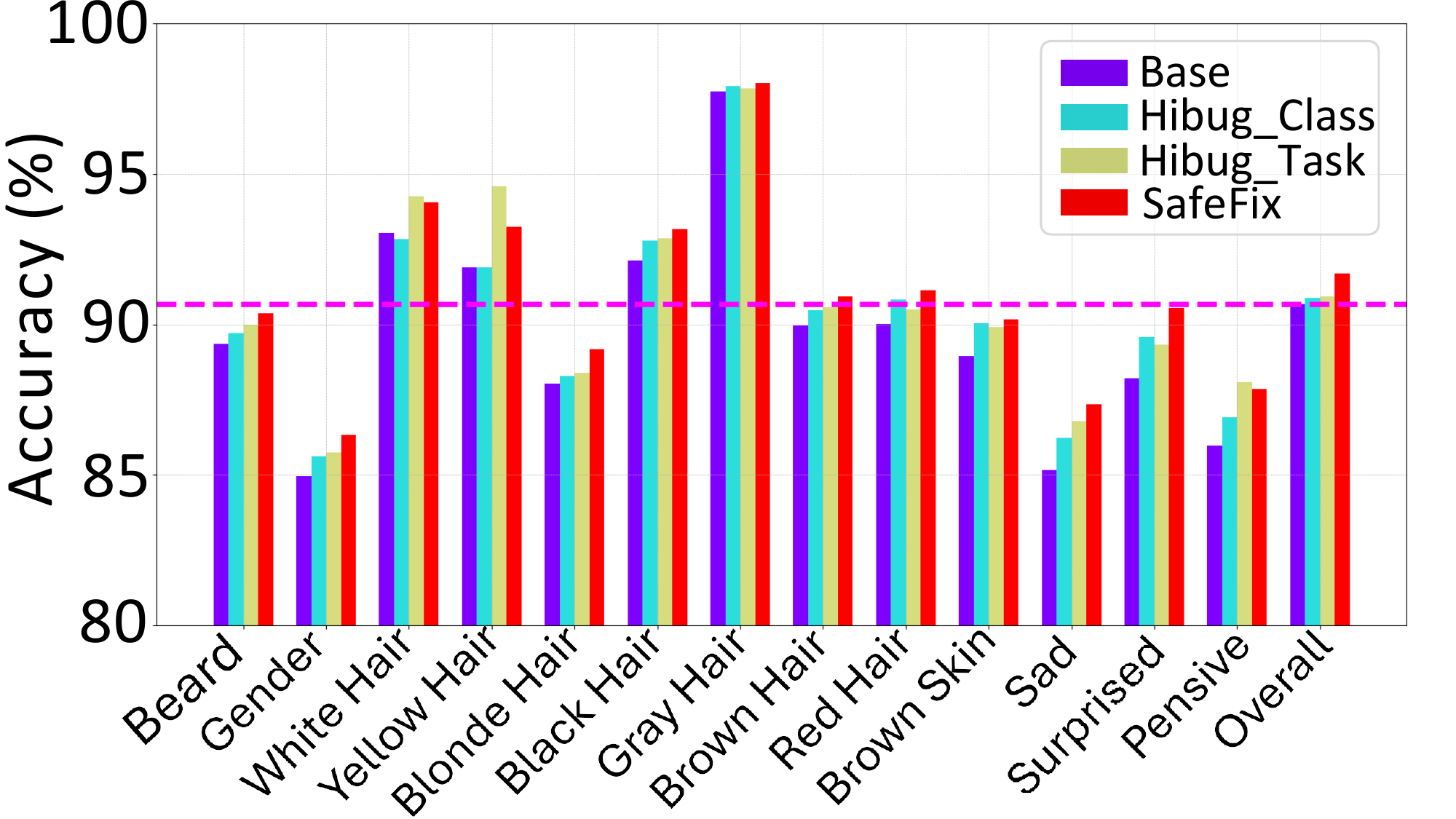}
    \caption{CelebA}
    \label{fig:fix_bugs_a}
  \end{subfigure}
  \hfill
  \begin{subfigure}[b]{0.49\linewidth}
    \includegraphics[width=\linewidth]{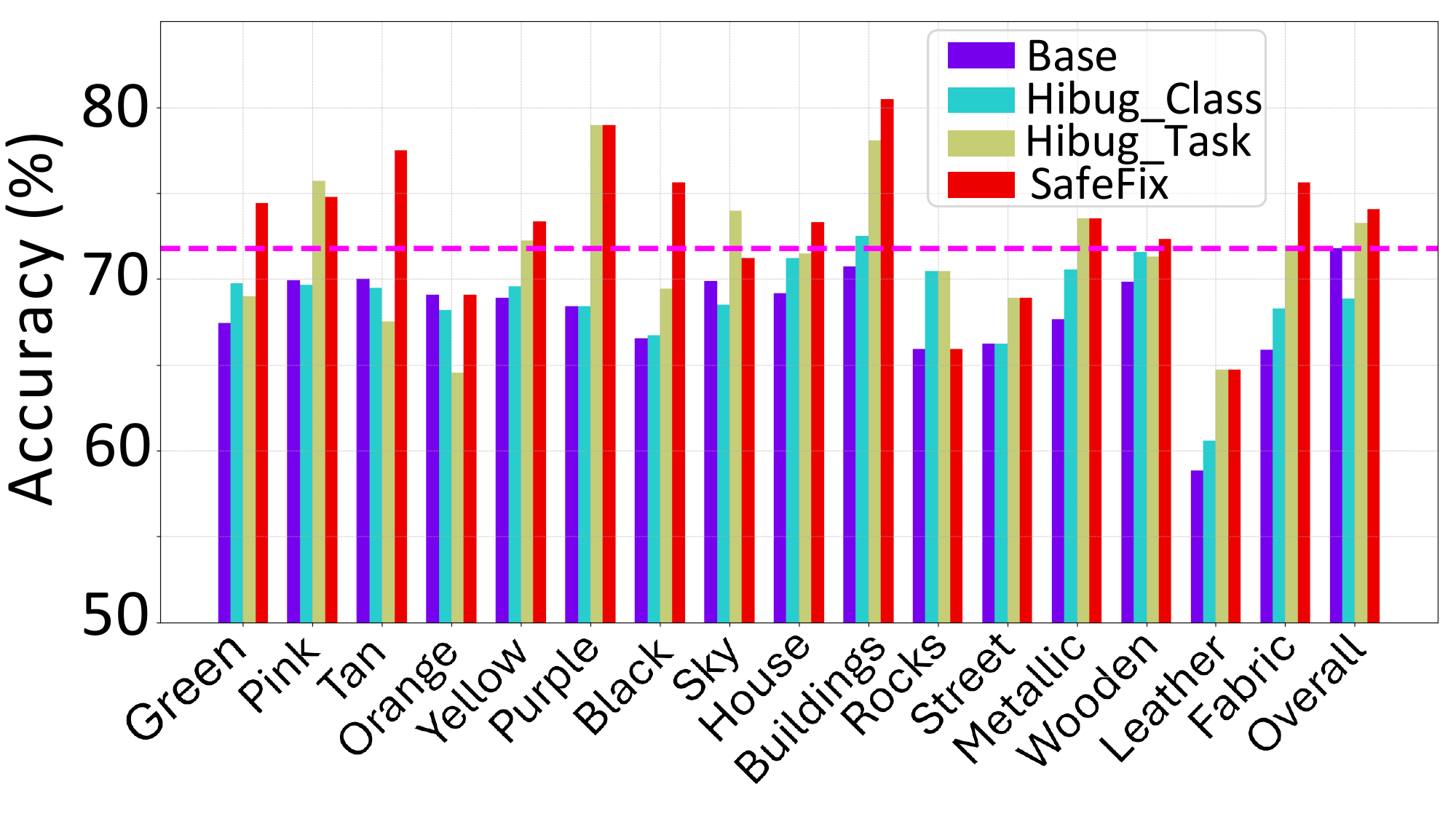}
    \caption{ImageNet10}
    \label{fig:fix_bugs_b}
  \end{subfigure}
  \vspace{-5pt}
  \caption{Accuracy comparison of ResNet-18 trained using different augmentation methods. The \textcolor{magenta}{dashed line} represents the average overall accuracy without additional synthetic training data.}
  \label{fig:fix_bugs}
  \vspace{-12pt}
\end{figure}

To verify that \model's improvements specifically address targeted rare-case bugs rather than merely enhancing overall performance, we analyze attribute-level validation accuracy changes on CelebA and ImageNet10, as shown in Figure~\ref{fig:fix_bugs}. For clarity, the ImageNet10 plot includes only three representative attributes—\texttt{color}, \texttt{background}, and \texttt{texture}—as the dataset contains many attribute dimensions. Specifically, the left plot highlights improvements for CelebA after adding 5,000 synthetic images to the original 80,000 training samples. The right plot demonstrates accuracy gains on ImageNet10, achieved by adding 100 synthetic images to the original training set of 10,400 samples.

Taking ImageNet10 as an example, our targeted synthetic augmentation on \texttt{pink color} and \texttt{fabric texture} significantly improved accuracy for these selected rare-case attributes. Accuracy for the \texttt{pink color} attribute increased from 69.90\% to 74.76\%, surpassing the overall accuracy of 74.31\%. Similarly, accuracy for the \texttt{fabric texture} attribute improved from 65.85\% to 75.61\%, also exceeding the overall accuracy.
In contrast, attributes not explicitly targeted by augmentation, such as the \texttt{rocks} background, exhibited minimal or no improvement—its accuracy remained unchanged—highlighting that the gains from \model are concentrated on the intended rare-case bugs rather than uniformly distributed across all attributes.

These substantial attribute-specific improvements confirm that \model effectively repairs identified rare-case failure slices rather than providing a generalized performance boost. \model also shows that all attributes improve across both datasets and introduces \textbf{no new bugs}, which indicates that the method remains stable outside the targeted regions. Attributes not targeted by augmentation show  negligible accuracy changes, further reinforcing that \model precisely and safely addresses the \textbf{targeted} rare-case failures.

\subsection{\model Directly Addresses Diagnosed Failure Modes, Not Merely Augments Data}

To verify that SafeFix’s performance gains stem from accurately fixing rare-case bugs rather than from generic data augmentation, we compare red-hair and yellow-hair augmentations on CLIP. As discussed in Section~\ref{sec:main_results}, we select red hair for ResNet and yellow hair for CLIP in the CelebA dataset, based on which attribute is more likely to trigger rare-case bugs. For CLIP, both \texttt{red hair} and \texttt{yellow hair} are low-frequency attributes that meet the rarity criterion in Eq.~\eqref{eq:rare}. However, only the \texttt{yellow hair} slice additionally  satisfies the low-accuracy criterion in Eq.~\eqref{eq:low_acc_and_rare}, making it a true rare-case bug slice for CLIP. 

\begin{figure}
  \vspace{-16pt}
  \centering
  \includegraphics[width=0.4\textwidth]{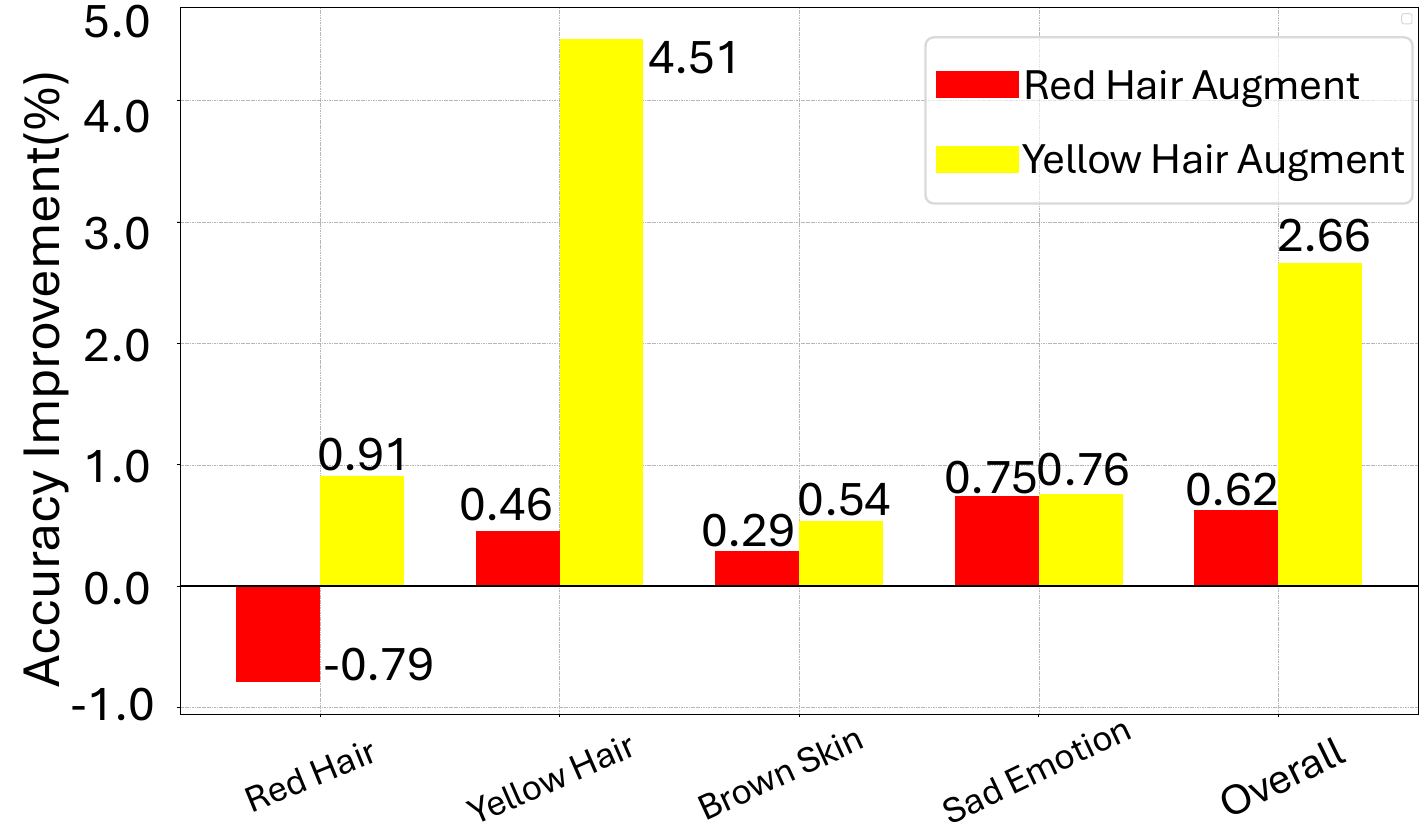}
  \vspace{-5pt}
  \caption{Test accuracy(\%) improvements for red-hair vs. yellow-hair augmentation on CLIP.}
  \label{fig:red_yellow_clip}
  \vspace{-14pt}
\end{figure}

Figure~\ref{fig:red_yellow_clip} confirms that augmenting 1000 images using yellow-hair samples—the diagnosed failure mode for CLIP—substantially improves accuracy on its target slice (\textbf{+4.51\%}). Furthermore, this targeted augmentation provides positive gains across all other attributes, including on the "Red Hair" slice (+0.91\%) and a significant boost to \textbf{"Overall" accuracy (+2.66\%)}. 
Conversely, augmenting with red hair, which is not a diagnosed bug, reveals a harmful outcome. Most notably, it degrades performance on its own target slice by \textbf{-0.79\%}. This counter-intuitive result contrasts with the positive gains seen for ResNet (Figure~\ref{fig:fix_bugs_a}), highlighting that different architectures can react to synthetic data in unpredictable ways. 
While red-hair augmentation does provide minor gains on other attributes (e.g., +0.75\% on \texttt{sad emotion}), its minimal overall accuracy impact (+0.62\%) and significant negative side effect on the targeted class itself underscore the importance of our diagnosis-driven approach. \textbf{Augmenting the \textit{correctly diagnosed} bug} (\texttt{yellow hair}) leads to \textbf{effective and safe repair}, while augmenting an \textit{undiagnosed} attribute (\texttt{red hair}) can be ineffective and harmful.

\subsection{Effect of Alternative Bug Slice Selections}

\begin{table}[t]
\centering
\small
\resizebox{.99\linewidth}{!}{%
\begin{tabular}{|r|cccccc|} 
\toprule
\cmidrule(lr){1-7}
\# images & RH & BS & SE & RH\_BS & RH\_SE & Ours \\ 
\midrule
1,000  & 7.85 & 8.59 & 11.77 & 7.74 & 10.39 & \textbf{14.32} \\
5,000  & 10.82 & 12.62 & 11.77 & \textbf{14.42} & 6.57 & 12.69 \\
10,000 & 11.03 & 10.29 & 12.62 & 9.97 & 9.33 & \textbf{14.95} \\
\bottomrule
\end{tabular}%
} 
\caption{Fix Rate (FR \%) for attribute-variant selections. Ours is the RH\_BS\_SE combination.}
\vspace{-14pt}
\label{tab:base_and_attribute_variants} 
\end{table}

Table~\ref{tab:base_and_attribute_variants} reports FR on CelebA using ResNet augmenting with different subsets of rare-case attributes. \textbf{RH}, \textbf{BS}, and \textbf{SE} refer to \texttt{red hair}, \texttt{brown skin}, and \texttt{sad emotion}, respectively. Combined settings like \textbf{RH\_BS} and \textbf{RH\_BS\_SE} (Ours) augment images with multiple attributes. Augmentation improves over the base model (90.57\%) across all settings. Notably, our three-way combination, \textbf{RH\_BS\_SE}, achieves the highest FR at $1,000$ and $10,000$ images and remains competitive at $5,000$ images, showing that targeting intersecting rare-case conditions is more effective than augmenting isolated attributes alone. Although \textbf{RH\_BS} alone achieves the highest FR at $5,000$ images, augmenting only that slice does not address concurrent failure modes, indicating persistent errors. Firstly, our \textbf{RH\_BS\_SE} combination improves all three slices concurrently (Figure~\ref{fig:fix_bugs_a}). Secondly, for the \textbf{RH\_BS\_SE} attribute combination, the number of bugs decreases from $10$ to $4$, while augmenting only \textbf{SE} or \textbf{RH\_BS} reduces the count to $7$ and $6$ respectively (using $\rho=0.05$ and $\epsilon=0.03$). A similar pattern holds on ImageNet10: selecting the \texttt{pink color} + \texttt{fabric texture} slice reduces rare-case bugs from $13$ to $7$, whereas augmenting only \texttt{pink color} reduces them to $9$ and only \texttt{fabric texture} reduces them to $10$.

After further verification, this reduction does not introduce any new bugs that were not present before, indicating that SafeFix repairs existing failures without adding new bugs on both datasets. The number of identified bugs is sensitive to these thresholds and using different threshold values can change the number of bugs fixed, as shown in Appendix~\ref{sec:threshold}. We note that larger attribute combinations (4 to 7 attributes) also operate correctly but reach lower FR (around 9\% on CelebA and 7\% on ImageNet10), falling below our selected attribute groups.

\subsection{Effect of the LVLM Filter and Human Audit}
\label{sec:human_audit}

\begin{wraptable}{r}{0.4\linewidth}
\vspace{-1pt}
\centering
\setlength{\tabcolsep}{4pt}
\renewcommand{\arraystretch}{1.0}
\small
\resizebox{0.85\linewidth}{!}{
\begin{tabular}{lcc}
\toprule
\textbf{Error Type} & \textbf{Before} & \textbf{After} \\
\midrule
BG    & 7  & 2 \\
ATTR  & 24 & 3 \\
\bottomrule
\end{tabular}
}
\vspace{-8pt}
\caption{\small{Average error rates (\%) before and after LVLM filtering.}}
\label{tab:lvlm_filter}
\vspace{-10pt}
\end{wraptable}

We use an LVLM as the attribute filter, and a human-based test yields nearly identical results. In a human audit conducted with 5 AI graduate students on 300 CelebA images, the pass rates are 97\% for ``red hair'', 99\% for ``brown skin'', 78\% for ``sad emotion'', and 98\% for the original label, with an average overlap of about 95\% with \qwen{} and 93\% with \llava{}, showing that the LVLMs closely match human verification. The main errors arise from background color changes (\textbf{BG}) ($7\%$) and the edited image failing to express the target attribute (\textbf{ATTR}) ($24\%$); the LVLM fixes most of them, as shown in Table~\ref{tab:lvlm_filter}.

\subsection{Ablation Study}
\label{sec:ablation}

\begin{table}[t]
  \centering
  \setlength{\tabcolsep}{4pt}
  \renewcommand{\arraystretch}{1.0}
  \small
  \resizebox{.75\linewidth}{!}{
    \begin{tabular}{l cc ccc}
      \toprule
      & \multicolumn{2}{c}{\textbf{Components}}
      & \multicolumn{3}{c}{\textbf{CelebA Fix Rate (FR \%)}} \\
      \cmidrule(lr){2-3}\cmidrule(lr){4-6}
      & CDM & LVLM & 1{,}000 & 5{,}000 & 10{,}000 \\
      \midrule
      a) &    &            & 6.79 & 4.88 & 3.92 \\
      b) &    & \checkmark & 7.21 & 4.77 & 6.99 \\
      c) & \checkmark &    & 8.06 & 8.27 & 10.07 \\
      d) & \checkmark & \checkmark & \textbf{14.32} & \textbf{12.09} & \textbf{14.95} \\
      \bottomrule
    \end{tabular}
  }
  \caption{Ablation on the impact of CDM and LVLM components at different scales, reported as Fix Rate (FR \%). This highlights SafeFix’s robustness under both sparse and saturated augmentations.}
  \label{tab:ablate_csd_lvlm}
  \vspace{-14pt}
\end{table}

We ablate the contributions of the two core components in our pipeline: the Conditional Diffusion Model (CDM) and the LVLM filter. Table~\ref{tab:ablate_csd_lvlm} reports the fix rate on CelebA when using different combinations of these components across three augmentation scales. Note that configuration (a) corresponds to the baseline strategy from \hibug~\cite{chen2023hibug}. We observe several trends: adding either the LVLM (b) or the CDM (c) individually improves accuracy across all scales; \model combining both the CDM and the LVLM (d) yields the highest performance in all cases.

\section{Conclusion}

We presented an automated model repair pipeline for vision tasks that combines failure‐attribute diagnostics with targeted synthetic augmentation. We use a Conditional Diffusion Model (CDM) to generate attribute-preserving variants and apply LVLM-based filtering to ensure semantic correctness, thereby focusing augmentation on true rare-case failure slices. Experiments on CelebA and ImageNet10 with ResNet, ViT, and CLIP backbones show consistent gains in accuracy and reduced bugs on underrepresented subpopulations, outperforming CDM-based and \hibug{} baselines. Ablations confirm that the CDM and the LVLM contribute complementary benefits, highlighting the importance of targeted, validated augmentation for robust model repair.

\noindent\textbf{Limitations.}
Our method’s effectiveness is constrained by its components. Crucially, the pipeline can inherit biases from the diffusion model or the LVLM, potentially perpetuating the fairness issues if these components have demographic bias. Other limitations include the computational cost of generation and filtering, potential image artifacts, and a fixed attribute vocabulary that cannot address unmodeled failure modes. Future work will focus on mitigating inherited biases, improving efficiency, and dynamic attribute discovery.

\clearpage
{
    \small
    \bibliographystyle{ieeenat_fullname}
    \bibliography{main}
}


\clearpage
\appendix

\section*{\LARGE Appendix}



\section{Attribute Assignment Details}
\label{sec:attribute_assignment}

\noindent \textbf{CelebA.} Attribute assignments for the CelebA dataset are based on the official annotations provided by the dataset and follow the settings used in \hibug~\cite{chen2023hibug}.

\noindent \textbf{ImageNet10.} Since ImageNet10 does not provide attribute annotations, we defined several attribute categories. As an example, we describe three representative categories: \texttt{color}, \texttt{background}, and \texttt{texture}. Each category is associated with a predefined set of candidate values, natural language prompt templates, and corresponding visual question answering (VQA) queries directed to the BLIP model. The structure is outlined below:

\begin{quote}
\textbf{color} \\
\hspace*{1em} \textit{Values:} red, green, pink, brown, white, blue, tan, silver, orange, gray, maroon, \hspace*{1em} yellow, multicolored, purple, black \\
\hspace*{1em} \textit{Prompt:} ``A photo of a \#1 \#LABEL.'' \\
\hspace*{1em} \textit{Question:} ``What color is the main object?''

\textbf{background} \\
\hspace*{1em} \textit{Values:} sky, trees, inside a house, buildings, grass, rocks, bridge, water, wall, \hspace*{1em} house, street, wild, snow \\
\hspace*{1em} \textit{Prompt:} ``The background of this photo is \#1.'' \\
\hspace*{1em} \textit{Question:} ``What is the background of this photo?''

\textbf{texture} \\
\hspace*{1em} \textit{Values:} plastic, metallic, wooden, leather, fabric, ceramic, glass \\
\hspace*{1em} \textit{Prompt:} ``A photo of a \#1 \#LABEL.'' \\
\hspace*{1em} \textit{Question:} ``What texture is the main object?''
\end{quote}

These definitions are used throughout the pipeline to assign, generate, and validate attribute-specific variations for model debugging on ImageNet10.

\section{Conditional Diffusion Model Details}
\label{appendix:cdm}

We use \textit{Stable Diffusion v1.5} as our text-to-image diffusion backbone and apply \controlnet~\cite{zhang2023adding} to enable fine-grained attribute control. The ControlNet model is conditioned on soft edge maps extracted from original training images using the HED detector~\cite{xie2015holistically}, which preserves structural consistency during generation. 

For each rare-case slice, we construct an attribute-preserving prompt. The tokens are mapped from attribute identifiers using the following dictionary:
\begin{verbatim}
token_map = {
  'redhair': 'vibrant red hair',
  'brownskin': 'brown skin',
  'sademotion': 'sad emotion'
}

\end{verbatim}

A final prompt is created by joining the mapped attribute phrases. For example:

\begin{quote}
\texttt{a person with vibrant red hair and brown skin, in sad emotion, not change other previous color, high detail, natural lighting}
\end{quote}

This prompt emphasizes the target attributes while including an explicit instruction to \texttt{not change other previous color}, helping the conditional diffusion model preserve unrelated visual aspects of the original image.

We use the following additional prompts during generation:

\begin{itemize}
  \item Positive prompt: \texttt{best quality, extremely detailed}
  \item Negative prompt: \texttt{lowres, bad anatomy, bad hands}
\end{itemize}

All generated images are later filtered by a vision--language model to ensure semantic correctness before being added to the training set.

\section{Large Vision--Language Model (LVLM) Details}
\label{app:lvlm-details}

\subsection{LVLM Query Format}
For semantic filtering of generated images, we use two large vision--language models (LVLMs), specifically \qwen~\cite{bai2025qwen25vl} and \llava~\cite{liu2023improvedllava}, to verify whether each image accurately satisfies its intended attribute-value conditions.

\textbf{CelebA.} For each generated image, we query the LVLM with a set of natural language questions tailored to the attributes of interest. For example:

\begin{itemize}
  \item For the \texttt{lipstick} attribute:
  \begin{quote}
  \texttt{Is the person in this picture wearing lipstick?}
  \end{quote}

  \item For other attributes (e.g., hair color or skin tone):
  \begin{quote}
  \texttt{Does the person in this picture have brown skin?}
  \end{quote}
\end{itemize}

\textbf{ImageNet10.} For each image–attribute pair, we construct a set of verification questions as follows:

\begin{quote}
\texttt{Does the \{class name\} have \{value\} \{attribute\}?} \\
\texttt{Is there a \{class name\} in the picture?}
\end{quote}

All prompts share a common instruction:
\begin{quote}
\texttt{For each question, only answer with 1 (yes) or 0 (no). Provide answers separated by spaces.}
\end{quote}
These structured prompts ensure that the LVLM produces reliable binary outputs for filtering. Only images for which all responses are ``1'' for the selected attributes and that retain the same labels are kept for training.

\subsection{Additional Analysis for Filtering Attributes}
\label{qwen_vs_llava}
In practice, we observed that the \qwen model often exhibits uncertainty when verifying emotional expressions. For some images, the emotion (e.g., \texttt{sad}) is inherently ambiguous and not reliably recognizable even by humans. In such cases, the LVLM often fails to respond decisively with a ``1'' or ``0''. This leads to a relatively low filtering pass rate for emotion-related attributes. This ambiguity reflects the intrinsic difficulty of emotion recognition, especially in the absence of strong visual cues.

Empirically, the pass rate of \qwen for emotion filtering is approximately 80\%. In contrast, most other attributes (e.g., \texttt{color}, \texttt{skin}, \texttt{texture}) already yield filtering accuracies above 90\%. When keeping the original label unchanged, the filtering results are mostly correct, with over 98\% of the generated samples preserving the original label. When combining label consistency with two or three target attributes in the CelebA dataset, the overall filtering process achieves an average accuracy of approximately 70\%. For the ImageNet10 dataset, where only two attributes—\texttt{color} and \texttt{texture}—are modified while keeping the label unchanged, the average filtering accuracy slightly exceeds 80\%.

Nonetheless, when using \llava, the pass rate for emotion filtering exceeds 95\%, while the filtering accuracy for other attributes remains around 90\%. We hypothesize that emotional attributes are more difficult for conditional diffusion models to modify precisely, which in turn makes it harder for LVLMs to detect whether the intended change is present. We choose to use \qwen as our default LVLM throughout most experiments to ensure consistency, reproducibility, and stricter evaluation of attribute presence, especially under less ideal generation conditions.


\section{Additional Implementation Details} 
We conduct all experiments on an NVIDIA A100 GPU.
We take the CelebA “wearing lipstick” classification task as an example. We evaluate three backbone architectures for the classification task. ResNet-18~\cite{he2016deep}, ViT-B/16~\cite{dosovitskiy2020vit}, and CLIP (ViT-B/32)~\cite{radford2021learning} are all initialized with random weights. All vision models are trained using cross-entropy loss. Optimization is performed using Adam~\cite{kingma2015adam} with a learning rate of \(1\times10^{-4}\) and weight decay of \(1\times10^{-3}\). All experiments use a batch size of 64 and are trained for 20 epochs.

For rare-case bug discovery and augmentation, we take the CelebA “wearing lipstick” classification task as an example. We first use a computer vision model \(f_\theta\colon \mathcal{X}\to\mathcal{Y}\) to produce predictions and compute the overall test accuracy \(\mathrm{Acc}(\mathcal{D}_{\mathrm{val}})\). 
For our experiments on the CelebA dataset, we use 7 distinct hair color attributes. These color categories were initially proposed using ChatGPT (GPT-4 with vision~\cite{openai2023gpt4}) and subsequently assigned to the images by the BLIP~\cite{li2022blip} model. This assignment process followed the same setting as in~\cite{chen2023hibug}. Accordingly, we set the rare threshold \(\rho = 0.05\) (which is less than \(1/7 \approx 0.143\)) and the accuracy difference threshold \(\epsilon = 0.03\). We found there are 2{,}484 red-hair images out of the 80{,}000 training samples, which corresponds to only 3.11\% of the data. Their slice accuracy is 0.8901, which is lower than the overall \(\mathrm{Acc}(\mathcal{D}_{\mathrm{val}}) = 0.9068\). Based on these thresholds, we identify the subset of red-hair images that satisfy both conditions and treat them as a candidate rare-case bug slice.
For rare-case bug discovery and augmentation, we follow the procedure described above, using accuracy and slice support to identify candidate attributes. We set the rarity threshold \(\rho = 0.05\) and the accuracy difference threshold \(\epsilon = 0.03\) based on dataset statistics. For synthetic augmentation, we use the \controlnet~\cite{zhang2023adding} as the conditional diffusion model, which is based on Stable Diffusion 1.5~\cite{rombach2022high} and conditioned on soft HED boundaries~\cite{xie2015holistically}. The soft HED boundary preserves fine structural details from the input images, making this approach particularly suitable for attribute-preserving edits such as recoloring and stylizing. To filter and validate generated images, we employ the \qwen~\cite{bai2025qwen25vl} as the primary large vision--language model (LVLM), and additionally compare results with \llava~\cite{liu2023improvedllava}.

In terms of computational cost, generating 1,000 images with ControlNet requires slightly over one hour, while the subsequent filtering of these images with our LVLM takes approximately 10 minutes. From an algorithmic complexity perspective, the ControlNet generation process exhibits a computational complexity of $O(D \times S \times U)$, where $D$ represents the number of images in the dataset, $S$ denotes the number of DDIM sampling steps (in our experiments we set $S=30$), and $U$ corresponds to the complexity of each UNet forward pass. For Stable Diffusion 1.5, each UNet forward pass requires approximately (\(\approx\)2.3 TFLOPs) of computation with 860M parameters operating on 64×64 latent representations. The dominant computational bottleneck lies in the iterative denoising process, where each of the 30 sampling steps requires a full forward pass through the UNet architecture. 

In contrast, our LVLM-based filtering pipeline demonstrates a more efficient complexity of $O(D \times I)$, where $I$ represents the inference time per image for the vision--language model. Empirical evaluation on the ImageNet images shows that our \llava{}-7B filtering achieves a throughput of 7,025 images per hour with an average processing time of 0.512 seconds per image. The significantly lower computational overhead of the filtering stage (approximately $7\times$ faster) stems from the single-pass nature of LVLM inference, eliminating the iterative sampling required by diffusion models. Resource utilization analysis reveals that the filtering process maintains stable GPU memory usage at 14.7GB with 98.6\% of computation time dedicated to model inference, demonstrating high computational efficiency. The filtering pipeline also achieves a 70-80\% success rate in identifying images matching the specified attributes, validating both the effectiveness of our multi-attribute prompting strategy and the quality of the dataset for the target criteria. This efficiency advantage becomes increasingly pronounced when processing large-scale datasets, as the filtering complexity scales linearly with dataset size while maintaining constant per-image processing time.

\section{Accuracy Results}
\label{sec:appendix_acc}

In the Results section, we introduce the Fix Rate (FR), which can be inferred from accuracy metrics. Therefore, we include the accuracy results in Table~\ref{tab:celeba_results} and Table~\ref{tab:imagenet_results} for the CelebA and ImageNet10 datasets, respectively.

On \textbf{CelebA}, we select rare-case bugs defined by attribute–value combinations \texttt{red hair}, \texttt{brown skin}, and \texttt{sad emotion} for ResNet and ViT, and \texttt{yellow hair}, \texttt{brown skin}, and \texttt{sad emotion} for CLIP, based on the most frequent patterns identified among failure slices. \model{} consistently achieves the highest test accuracy across all models (ResNet, ViT, and CLIP) and varying levels of synthetic augmentation as shown in Table~\ref{tab:celeba_results}. For example, with 1{,}000 added images, our method improves accuracy by +1.35\% (ResNet), +2.89\% (ViT), and +2.78\% (CLIP) relative to their respective baselines. These results show that our attribute-targeted augmentation and filtering pipeline is effective in repairing rare-case failure slices, outperforming both CDM-based and \hibug{} baselines.

On \textbf{ImageNet10}, similar trends emerge, as shown in Table~\ref{tab:imagenet_results}. For ResNet and ViT, we target rare-case bugs involving \texttt{pink color} and \texttt{fabric texture}, while for CLIP we use \texttt{orange color} and \texttt{fabric texture}. Across all models, our proposed method consistently surpasses the baseline methods. Specifically, our method improves ResNet accuracy by +3.18\% (at 100 images) compared to the base model. ViT and CLIP also exhibit a steady improvement compared to other methods.

\begin{table*}[!t]
\centering
\resizebox{1.0\textwidth}{!}{%
{
\small
\setlength{\tabcolsep}{4pt}
\renewcommand{\arraystretch}{0.95}

\begin{tabular}{l | C{1.44cm} C{1.44cm} C{1.44cm} | C{1.44cm} C{1.44cm} C{1.44cm} | C{1.44cm} C{1.44cm} C{1.44cm}}
\toprule
\multirow{2}{*}{Method} 
  & \multicolumn{3}{c|}{\textbf{ResNet}} 
  & \multicolumn{3}{c|}{\textbf{ViT}} 
  & \multicolumn{3}{c}{\textbf{CLIP}} \\
\cmidrule(lr){2-4} \cmidrule(lr){5-7} \cmidrule(lr){8-10}
  & 1k Images & 5k Images & 10k Images
  & 1k Images & 5k Images & 10k Images
  & 1k Images & 5k Images & 10k Images \\
\midrule
Base               & \multicolumn{3}{c|}{90.57} & \multicolumn{3}{c|}{85.02} & \multicolumn{3}{c}{88.32} \\
\cmidrule(lr){1-10}
DA-CDM             & 90.95 & 90.88 & 90.86 & 85.63 & 86.55 & 86.04 & 90.11 & 90.25 & 90.14 \\
Mask-ControlNet    & 91.08 & 91.26 & 91.15 & 86.71 & 86.59 & 86.95 & 90.20 & 90.18 & 90.28 \\
HiBug\_Class       & 91.05 & 91.12 & 90.89 & 85.79 & 86.18 & 86.10 & 90.03 & 90.07 & 89.95 \\
HiBug\_Task        & 91.21 & 91.03 & 90.94 & 87.82 & 86.78 & 86.18 & 90.16 & 90.09 & 90.03 \\
\cmidrule(lr){1-10}
\textbf{Ours (L)} & 91.55 & 91.65 & 91.67 & 87.73 & \textbf{87.78} & 87.32 & 90.92 & 90.95 & \textbf{90.86} \\
\textbf{Ours (Q)} & \textbf{91.92} & \textbf{91.71} & \textbf{91.98} & \textbf{87.91} & 87.33 & \textbf{87.41} & \textbf{90.98} & \textbf{91.02} & 90.71 \\
\bottomrule
\end{tabular}
}%
}
\caption{Test accuracy (\%) on CelebA for varying numbers of added images, models, and methods. Ours (L) and Ours (Q) denote using \llava{}-7B and \qwen{}-7B as the large vision-language model filter, respectively.} 
\label{tab:celeba_results}
\end{table*}

\begin{table*}[!t]
\centering
\small
\setlength{\tabcolsep}{4pt}
\renewcommand{\arraystretch}{0.95}

\resizebox{1\linewidth}{!}{
\begin{tabular}{l | C{1.44cm} C{1.44cm} C{1.44cm} | C{1.44cm} C{1.44cm} C{1.44cm} | C{1.44cm} C{1.44cm} C{1.44cm}}
\toprule
\multirow{2}{*}{Method} 
  & \multicolumn{3}{c|}{\textbf{ResNet}} 
  & \multicolumn{3}{c|}{\textbf{ViT}} 
  & \multicolumn{3}{c}{\textbf{CLIP}} \\
\cmidrule(lr){2-4} \cmidrule(lr){5-7} \cmidrule(lr){8-10}
  & 100 Images & 500 Images & 1k Images
  & 100 Images & 500 Images & 1k Images
  & 100 Images & 500 Images & 1k Images \\
\midrule
Base               & \multicolumn{3}{c|}{71.73} & \multicolumn{3}{c|}{97.42} & \multicolumn{3}{c}{93.78} \\
\cmidrule(lr){1-10}
DA-CDM             & 73.21 & 73.80 & 74.14 & 97.58 & 97.76 & 97.77 & 93.96 & 94.05 & 94.15 \\
Mask-ControlNet    & 73.22 & 72.54 & 72.39 & 97.25 & 97.69 & 97.52 & 93.44 & 93.88 & 93.70 \\
HiBug\_Class       & 72.24 & 72.15 & 71.94 & 97.12 & 97.07 & 97.28 & 93.92 & 94.21 & 93.83 \\
HiBug\_Task        & 73.51 & 72.88 & 73.36 & 97.66 & 97.78 & 97.58 & 93.29 & 94.18 & 94.15 \\
\cmidrule(lr){1-10}
\textbf{Ours (L)}  & 74.10 & \textbf{74.19} & 73.92 & 98.21 & \textbf{98.17} & 98.11 & 94.49 & 94.80 & 94.44 \\
\textbf{Ours (Q)}  & \textbf{74.31} & 73.43 & \textbf{74.88} & \textbf{98.22} & 98.09 & \textbf{98.41} & \textbf{94.57} & \textbf{94.88} & \textbf{94.95} \\
\bottomrule
\end{tabular}
}
\caption{Test accuracy (\%) on ImageNet10 under varying image counts, models, and methods.}
\label{tab:imagenet_results}
\vspace{-14pt}
\end{table*}

\section{Large-Scale Settings in Main Results}
\label{sec:largescale}

In the main experiments,
 we show that \model improves test accuracy by augmenting underrepresented slices. To further assess its effectiveness under large-scale augmentation, we consider a scenario where a previously rare-case attribute—\texttt{red hair}—becomes common in training after adding 10{,}000 generated images (around 12{,}480 red-hair images out of 90{,}000 total), while its prevalence in the fixed validation set remains low (621 out of 20{,}000).

This setup creates an important distinction: although \texttt{red hair} is no longer rare in training (\(\sim14\%\)), it remains rare in validation (\(\sim3\%\)). Nevertheless, the accuracy on the red-hair subset improves significantly, demonstrating that \model effectively improves generalization to previously underperforming groups.

\paragraph{Interpretation.} This setup aligns with the goal of rare-case debugging: the term “rare case” refers to slices that are rare and inaccurate in evaluation or real-world deployment. The objective is to increase their representation in training to address model failure. Even though \texttt{red hair} is no longer rare in training, the model successfully reduces its failure rate in evaluation—without changing the validation distribution.

Our fixed validation set continues to reflect real-world statistics, ensuring that improvements in rare-case performance are meaningful. This practice—augmenting rare cases in training while keeping the test distribution fixed—is consistent with standard methodology in prior work~\cite{chen2023hibug}. These findings confirm that \model improves generalization on rare-case bugs without sacrificing reliability or introducing new biases.

\section{Additional Experiments}
\label{sec:additional-experiments}

\subsection{Generation Quality Analysis}
We compare \hibug{} and \model{} using four key metrics—LPIPS diversity~\cite{zhang2018unreasonable}, CLIP consistency~\cite{radford2021learning}, Fréchet Inception Distance (FID)~\cite{heusel2017gans} against real images, and the KL divergence of deep-feature distributions—using the ImageNet10 dataset with the \texttt{pink color} + \texttt{fabric texture} setting as an illustrative example, as shown in Table~\ref{tab:gen_metrics}.

\begin{table}[ht]
  \centering
  \small
  \setlength{\tabcolsep}{8pt}
  \renewcommand{\arraystretch}{1.1}
  \begin{tabular}{lcc}
    \toprule
    \textbf{Metric} & \textbf{HiBug\_Task}& \textbf{SafeFix} \\
    \midrule
    LPIPS diversity $\downarrow$    & 0.7582           & \textbf{0.6764}   \\
    CLIP consistency $\uparrow$   & \textbf{0.2087}           & 0.2049   \\
    FID (Real→Gen) $\downarrow$      & 0.2405           & \textbf{0.1490}   \\
    KL divergence $\downarrow$ ($\times10^{6}$) & 7.05               & \textbf{5.71}   \\
    \bottomrule
  \end{tabular}
  \caption{Generation‐quality metrics for \hibug{} vs.\ \model{} on ImageNet10 (pink‐color + fabric‐texture). “Real→Gen” indicates the Fréchet distance computed between the real and generated image feature distributions; lower FID means the generated set is closer to the real distribution. Arrows indicate preference: $\uparrow$ higher is better; $\downarrow$ lower is better.}
  \label{tab:gen_metrics}
\end{table}

\model{} achieves substantially lower FID and KL divergence compared to \hibug{}, indicating that its samples lie much closer to the real‐image distribution. Although its LPIPS diversity is moderately reduced and CLIP consistency slightly lower, these modest trade‐offs yield more reliable, attribute‐faithful examples for downstream repair. In practice, fewer but higher‐quality variants help correct rare-case failures without introducing noisy or out‐of‐distribution artifacts.

\subsection{SafeFix with alternative conditional diffusion models}

The targeted generation stage in our pipeline uses ControlNet\cite{zhang2023adding} as the conditional diffusion model. Although ControlNet is effective due to its stable structure-preserving behavior, relying on a single generator raises questions about whether SafeFix depends on ControlNet-specific features. To test the generality of the pipeline, we replaced ControlNet with another conditional diffusion model, InstructPix2Pix~\cite{brooks2023instructpix2pix}, while keeping the LVLM filter (\qwen{}). The resulting accuracy on CelebA, shown in Table~\ref{tab:instructpix2pix_vs_controlnet}, exceeds all baselines reported in the main results and is only slightly lower than SafeFix with ControlNet. This demonstrates that the pipeline is not tied to ControlNet and functions well with alternative generators.
\begin{table}[!t]
\centering
\small
\setlength{\tabcolsep}{4pt}
\renewcommand{\arraystretch}{0.95}

\begin{tabular}{l | C{1.44cm} C{1.44cm} C{1.44cm}}
\toprule
\multirow{2}{*}{Method} 
  & \multicolumn{3}{c}{\textbf{ResNet} (base acc: 90.57\%)} \\
\cmidrule(lr){2-4}
  & 1k Images & 5k Images & 10k Images \\
\midrule
DA-CDM           & 4.03 & 3.29 & 3.08 \\
Mask-ControlNet  & 5.41 & 7.32 & 6.15 \\
HiBug\_Class     & 5.09 & 5.83 & 3.40 \\
HiBug\_Task      & 6.79 & 4.88 & 3.92 \\
\cmidrule(lr){1-4}
\textbf{Ours (InstructPix2Pix)} 
                 & 11.32 & 11.85 & 11.74 \\
\textbf{Ours (ControlNet)} 
                 & \textbf{14.32} & \textbf{12.09} & \textbf{14.95} \\
\bottomrule
\end{tabular}
\caption{Fix Rate (FR\%) using different conditional diffusion models in the targeted generation stage. InstructPix2Pix~\cite{brooks2023instructpix2pix} serves as an alternative generator and shows strong repair performance.}
\label{tab:instructpix2pix_vs_controlnet}
\end{table}

\subsection{Qualitative Analysis of Attribute Editing and Failure Cases}
\label{sec:qualitative-attribute-edit}

\begin{figure}[!t]
  \centering
  \begin{subfigure}[b]{0.48\textwidth}
    \centering
    \includegraphics[width=\textwidth]{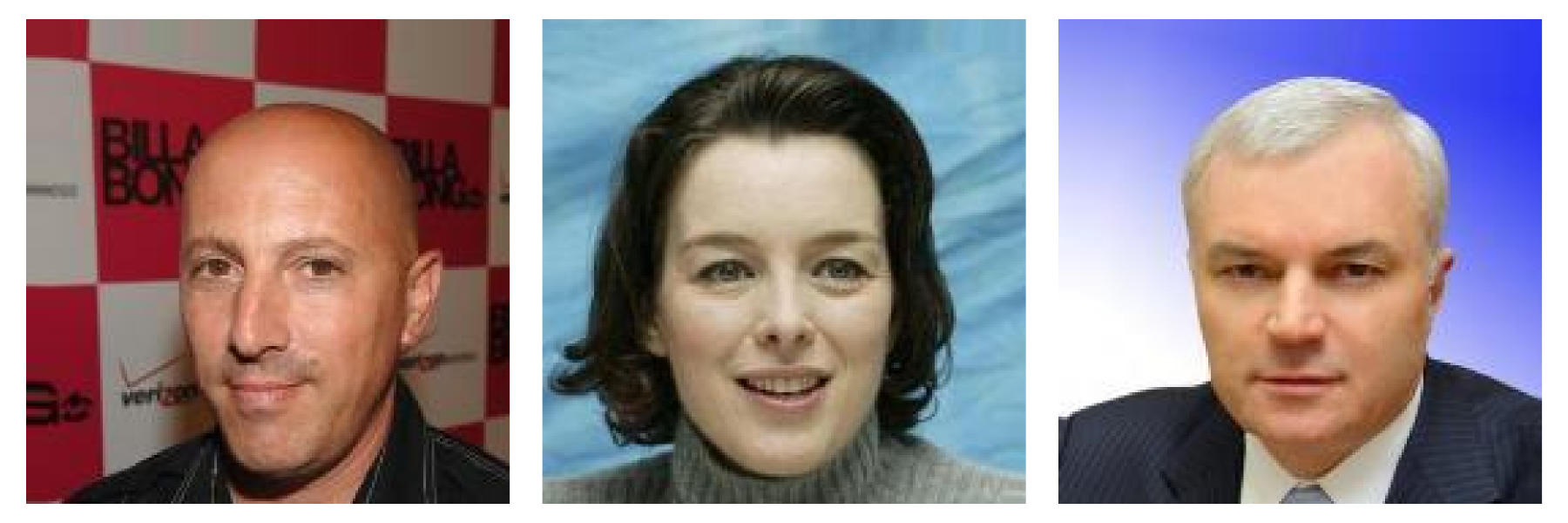}
    \caption{Original CelebA samples}
    \label{fig:celeba-original}
  \end{subfigure}
  \hfill
  \begin{subfigure}[b]{0.48\textwidth}
    \centering
    \includegraphics[width=\textwidth]{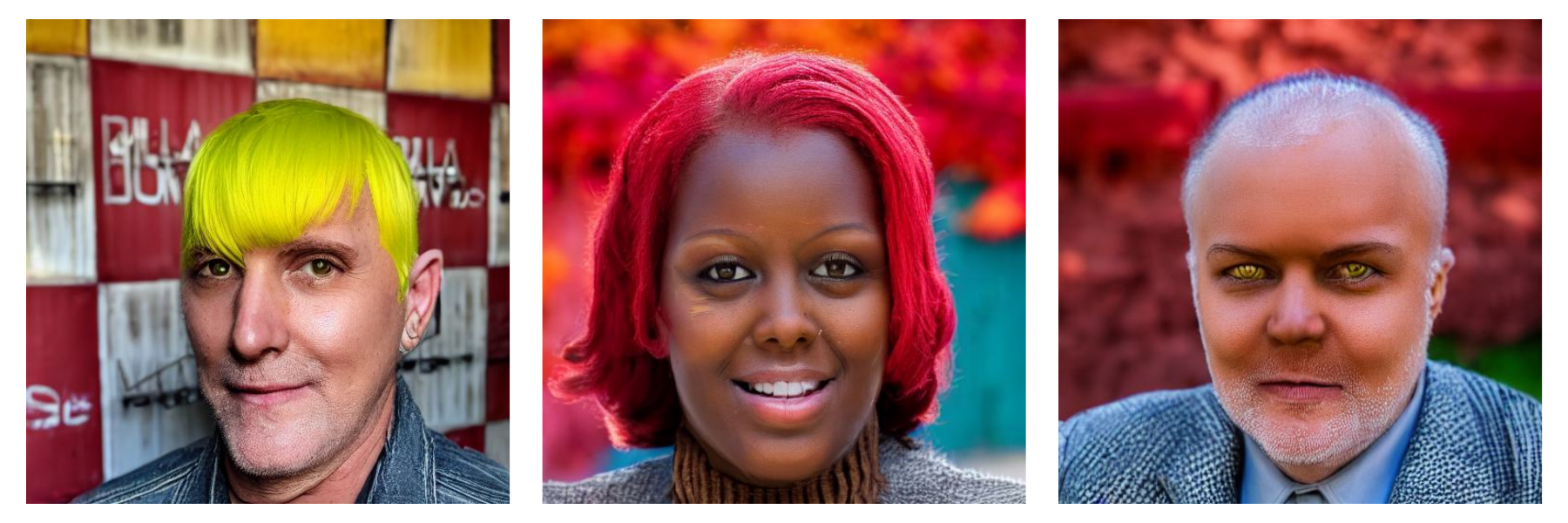}
    \caption{Attribute-modified images from \model}
    \label{fig:celeba-ours}
  \end{subfigure}
  \caption{Comparison between original CelebA images (left) and attribute-edited outputs (right) generated by \model.}
  \label{fig:grid-celeba-comparison}
\end{figure}

Figure \ref{fig:picture_compare} shows some good examples generated by the conditional diffusion model, but conditional diffusion models can make mistakes. In the CelebA dataset, we primarily target rare-case attributes such as red hair (RH), brown skin (BS), and sad emotion (SE), but our framework is flexible and supports other attribute configurations. Figure~\ref{fig:grid-celeba-comparison} shows examples of original and \model-generated images. In the first example, only yellow hair was added to the original face while all other identity features remained unchanged. The second image reflects a transformation into the red hair and brown skin attribute combination. In the third example, the model aimed to edit the hair color to red, but instead altered the background to a reddish hue without changing the hair itself—highlighting a typical failure of generative models when not filtered.

\begin{figure}[!t]
  \centering
  \begin{subfigure}[b]{0.48\textwidth}
    \centering
    \includegraphics[width=\textwidth]{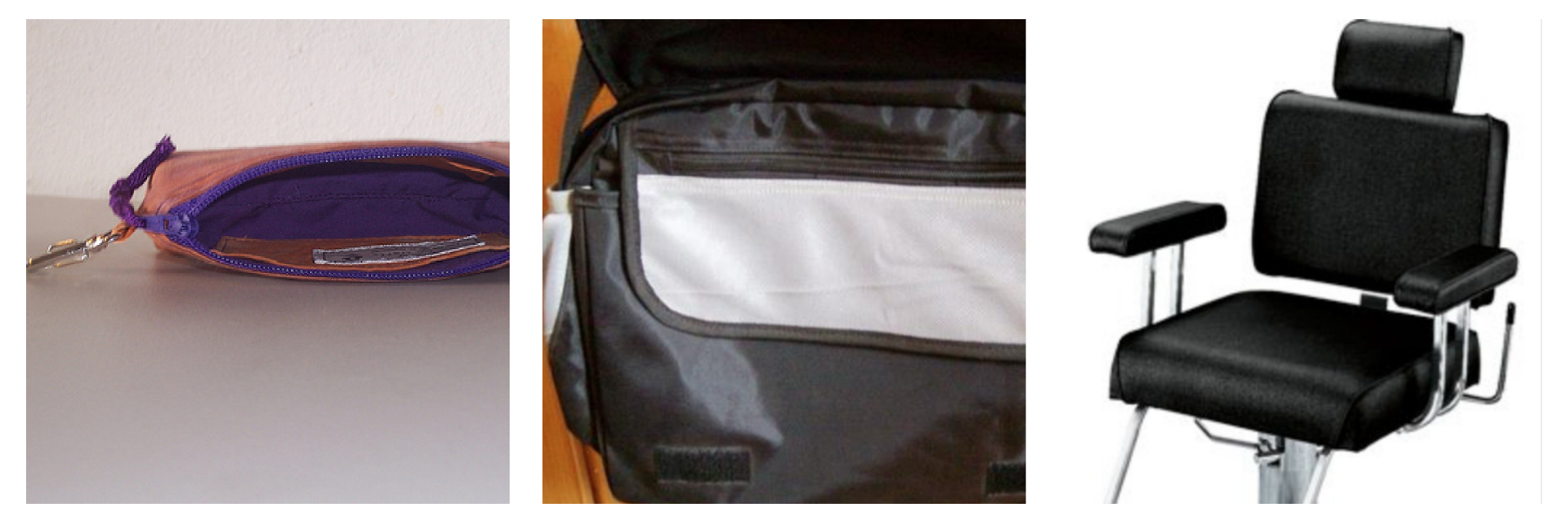}
    \caption{Original ImageNet10 samples}
    \label{fig:imagenet-original}
  \end{subfigure}
  \hfill
  \begin{subfigure}[b]{0.48\textwidth}
    \centering
    \includegraphics[width=\textwidth]{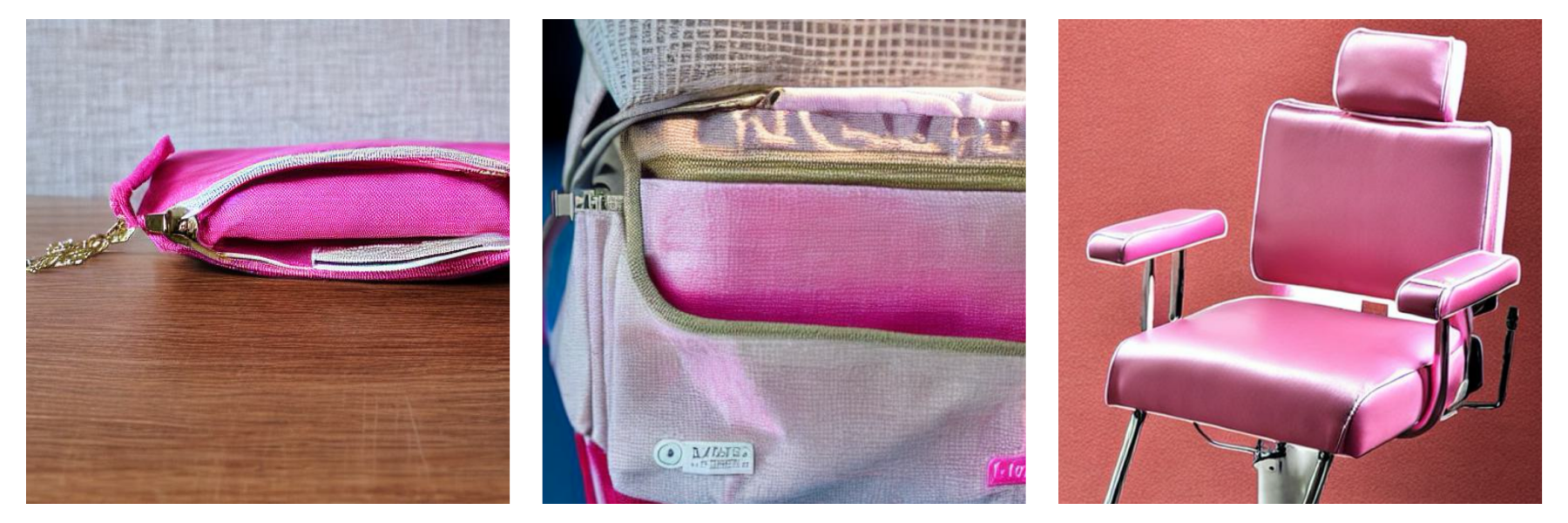}
    \caption{Attribute-modified images from \model}
    \label{fig:imagenet-ours}
  \end{subfigure}
  \caption{Comparison between original ImageNet10 images (left) and attribute-edited outputs (right) by \model targeting pink color and fabric texture.}
  \label{fig:grid-imagenet-comparison}
\end{figure}

Figure~\ref{fig:grid-imagenet-comparison} shows attribute editing results on ImageNet10 for the pink color and fabric texture combination. In the first example, a multicolored leather purse is successfully transformed into a pink fabric purse. The second example shows a black plastic backpack converted into a pink fabric version, though the modified velcro strap appears slightly unnatural—suggesting semantic correctness but imperfect realism. In the third case, a black leather barber chair becomes pink, but its texture remains visibly leather-like, indicating that not all attributes were successfully edited.

While over 90\% of generated images (as verified by \qwen) accurately reflect the target attributes, occasional failures illustrate that conditional diffusion models (CDMs) can still make subtle mistakes. These examples are representative failure cases selected for analysis; most generated images do not exhibit such issues. The majority of CelebA and ImageNet10 samples are correctly modified, but these rare errors underscore the importance of LVLM-based filtering to ensure semantic correctness and support reliable model repair.


\subsection{Evaluating Synthetic Data as a Standalone Training Set}
\label{sec:synthetic_standalone} 

\begin{table}[h] 
\centering
\setlength{\tabcolsep}{6pt}
\renewcommand{\arraystretch}{1.1}
\small
\begin{tabular}{rcc}
\toprule
\# Images & HiBug\_Task & \textbf{Ours} \\
\midrule
1{,}000  & 63.81 & \textbf{81.95} \\
5{,}000  & 76.44 & \textbf{90.54} \\
10{,}000 & 83.03 & \textbf{90.87} \\
\bottomrule
\end{tabular}
\caption{Test accuracy (\%) on $N$ synthetic images, trained from scratch. The $N$ images were split into an 8:1:1 train--val--test partition for this experiment.}
\label{tab:valsplit_retrain} 
\vspace{-1em}
\end{table}

In our main experiments, we added $N$ validated synthetic images to $\mathcal{D}_{\mathrm{train}}$ while keeping $\mathcal{D}_{\mathrm{val}}$ and $\mathcal{D}_{\mathrm{test}}$ fixed. To further examine the end-to-end value of our augmentation, we instead randomly split the $N$ synthetic images into an 8:1:1 train--val--test partition and trained ResNet from scratch. As shown in Table~\ref{tab:valsplit_retrain}, our method consistently outperforms \hibug{} by large margins across all settings. This suggests that \model, using a conditional diffusion model, generates images with more internally coherent distributions—that is, the synthetic samples are more semantically consistent with each other. It also shows that LVLM filtering reliably ensures label correctness, resulting in improved downstream classification accuracy. Together, these two key components—the CDM and the LVLM—underscore the semantic fidelity and effectiveness of our augmentation strategy.

\subsection{Threshold settings for identifying rare-case slices}
\label{sec:threshold}

Settings for the rare threshold ($\rho$) and the accuracy-difference threshold ($\epsilon$) are dataset-dependent. As shown in Table~\ref{tab:rho_threshold}, with $\rho=0.05$, the number of identified bugs drops from 10 to 4 on CelebA and from 13 to 7 on ImageNet10. For CelebA, increasing \(\rho\) to 0.06, 0.07, and 0.08 yields the same result, which raises the pre-fix bug count to 12, while the post-fix count remains stable at 4. Extreme values of $\rho$ become invalid: for ImageNet10, which contains 15 color categories, a value such as $\rho=0.2$ exceeds the $1/15$ ratio of each color and thus no longer reflects a rare case.

\begin{table}[!t]
\centering
\small
\setlength{\tabcolsep}{6pt}
\renewcommand{\arraystretch}{0.95}
\begin{tabular}{lccc}
\toprule
\textbf{Dataset} & \boldmath$\rho$ & \textbf{Pre} & \textbf{Post} \\
\midrule
CelebA      & 0.01        & 3  & 2 \\
CelebA      & 0.05        & 10 & 4 \\
CelebA      & 0.06  & 12 & 4 \\
CelebA      & 0.07  & 12 & 4 \\
CelebA      & 0.08  & 12 & 4 \\
CelebA      & 0.14        & 15 & 9 \\
ImageNet10  & 0.01        & 4  & 3 \\
ImageNet10  & 0.05        & 13 & 7 \\
ImageNet10  & 0.066       & 16 & 9 \\
\bottomrule
\end{tabular}
\caption{Effect of rare threshold $\rho$ on the number of identified bugs before and after repair.}
\label{tab:rho_threshold}
\end{table}

The accuracy-difference threshold $\epsilon$ follows a similar trend. Small values isolate a limited set of failure slices, while moderate values increase the number of identified bugs without affecting the repaired result. Table~\ref{tab:epsilon_threshold} shows that on both datasets, using \(\epsilon\) equals to 0.01, 00.02 and 0.03 yields stable post-fix counts. Extreme thresholds such as $\epsilon=0.1$ become unreliable because they classify many non-critical cases as failures, inflating both pre- and post-fix bug counts.

\begin{table}[!t]
\centering
\small
\setlength{\tabcolsep}{6pt}
\renewcommand{\arraystretch}{0.95}
\begin{tabular}{lccc}
\toprule
\textbf{Dataset} & \boldmath$\epsilon$ & \textbf{Pre} & \textbf{Post} \\
\midrule
CelebA      & 0.01 & 6  & 3 \\
CelebA      & 0.02 & 8  & 3 \\
CelebA      & 0.03 & 10 & 4 \\
CelebA      & 0.10 & 13 & 9 \\
ImageNet10  & 0.01 & 7  & 4 \\
ImageNet10  & 0.02 & 9  & 5 \\
ImageNet10  & 0.03 & 13 & 7 \\
ImageNet10  & 0.10 & 15 & 12 \\
\bottomrule
\end{tabular}
\caption{Effect of accuracy-difference threshold $\epsilon$ on the number of identified bugs before and after repair.}
\label{tab:epsilon_threshold}
\end{table}

\section{Future Work}
\label{sec:future_work}

\model{} demonstrates that \emph{targeted} synthetic augmentation plus LVLM
filtering can reliably repair rare–case bugs in image classification.  
Several promising research directions remain:

\begin{itemize}
  \item \textbf{Data–selection vs.\ data–generation.}  
        HiBug2~\cite{chen2025hibug2} focuses on \emph{discovering} failure slices and then
        \emph{retrieving} real images from the web to
        rebalance those slices instead of \emph{generating} images.  A natural extension is to combine the
        web–retrieval paradigm with SafeFix’s LVLM verifier—e.g.\ apply the
        same binary\,/\,attribute QA filter to crawled images before
        retraining. This may further reduce covariate shift without the
        risk of generative artifacts.


  \item \textbf{Closed–loop repair.}  
        We can propose an iterative ``detect\,→\,augment\,→\,retrain''
        loop.  Embedding \model{} inside a similar loop—re-running model
        diagnostics after each repair round—could automatically converge
        until no slice violates the rarity+error criteria.

  \item \textbf{Dynamic attribute discovery.}  
        Current diagnostics rely on a fixed attribute vocabulary.
        Future work can allow the LVLM to propose \emph{new} attributes on
        failure slices, making the repair pipeline fully open-ended.

  \item \textbf{Broader modalities and tasks.}  
        The same controlled-generation + LVLM filter idea is applicable to
        pose estimation, object detection, image captioning, and VQA.
        Extending \model{} to multi-modal inputs (e.g.\ video or text–image
        pairs) is an exciting avenue, especially for reasoning-centric
        benchmarks.
        
    \item \textbf{Beyond discrete attributes.} Discrete vocabularies of attributes are common failure sources, hence our focus. Future work can extend this direction by handling continuous or fine-grained attributes, supporting multi-attribute interactions, and studying how these repair signals transfer across datasets and model families.
\end{itemize}

\textbf{Societal Impact.} The proposed method could be important for real-world applications such as fairness auditing and reliability improvement in vision systems deployed in robotics, healthcare, surveillance, and auto-driving, by automatically identifying and repairing failure modes associated with underrepresented or biased visual patterns.

Overall, \model{} provides a foundation for a general \emph{data-centric}
loop: \textbf{diagnose $\rightarrow$ generate / retrieve $\rightarrow$
filter $\rightarrow$ retrain}.  Enhancing each stage with active
selection, dynamic attributes, and multi-task diffusion backbones
represents a rich future research agenda.


\end{document}

%% file: sec/0_abstract.tex
\begin{abstract}
Deep learning models for visual recognition often exhibit systematic errors due to underrepresented semantic subpopulations. Although existing debugging frameworks can pinpoint these failures by identifying key failure attributes, repairing the model effectively remains difficult. Current solutions often rely on manually designed prompts to generate synthetic training images—an approach prone to distribution shift and semantic errors. To overcome these challenges, we introduce a model repair module that builds on an interpretable failure attribution pipeline. Our approach uses a conditional diffusion text-to-image model to generate semantically faithful and targeted images for failure cases. To preserve the quality and relevance of the generated samples, we further employ a large vision--language model (LVLM) to filter the outputs, enforcing alignment with the original data distribution and maintaining semantic consistency. The effectiveness of this filtering step is verified by human audit. By retraining vision models with the synthetic dataset and source dataset, we significantly reduce errors associated with rare cases. Our experiments demonstrate that this targeted repair strategy improves model robustness without introducing new bugs. Code is available at \url{https://github.com/oxu2/SafeFix}. 
\end{abstract}